# RefGlass-GS: A UAV-Enabled Fusion Framework for Photorealistic, Semantic and Interactive Digitization of Reflective Glass Façades via Gaussian Splatting

Zhenyu Liang [ab], Xiao Zhang [a*], Boyu Wang [a], Zhaolun Liang [a], Ang Li [a], Jeff Chak Fu Chan [c], Mingzhu Wang [d], Jack C.P. Cheng [a*]

[a] Department of Civil and Environmental Engineering, The Hong Kong University of Science and Technology, Hong Kong, China
[b] Department of Civil and Environmental Engineering, University of Maryland, College Park, Maryland, USA
[c] Department of the Built Environment, National University of Singapore, Singapore
[d] Department of Architecture and Civil Engineering, City University of Hong Kong, Hong Kong, China
* Corresponding Authors
Email: zhenyu.liang@connect.ust.hk, xiao.zhang@connect.ust.hk, bwangbb@connect.ust.hk, lionel_a.li@connect.ust.hk, zliangaq@connect.ust.hk, chak.fu.chan@u.nus.edu, m.wang@cityu.edu.hk, cejcheng@ust.hk.

**Highlights:**
End-to-end architecture with real-world applications for glass façade digitization.
Individual glass panel segmentation using MAP estimation and structural regularities.
UAV view planning formulation ensures maximal coverage of view-dependent appearance.
Deferred shading with Reflection MLP to model high-frequency near-field reflections.
Standardized object-based GS data organization for interactive facility management.

**Abstract:**
Existing digitization of buildings with reflective glass façades suffers from geometric reconstruction distortion, unrealistic view-dependent texture rendering, and difficulties in object-based semantic enhancement. Therefore, we propose RefGlass-GS, a fusion framework that enables end-to-end UAV-based photorealistic, semantic, and interactive digitization of reflective glass façades. The contributions include: (1) proposing an individual glass panel segmentation method based on maximum a posteriori estimation with structural regularities, robust to severe reflection and background interference; (2) formulating a UAV viewpoint planning optimization function that maximizes the coverage of view-dependent appearance for sufficient data capture; (3) developing an optimized Gaussian Splatting framework with a Reflection MLP, a novel deferred shading function, and two enhanced regularization terms for effective modeling of high-frequency near-field reflections; (4) introducing a standardized data organization paradigm for structuring GS-based representations into object-based models, facilitating interactive facility management on digital twin platforms. Experiments on real-world reflective glass façade scenes validate the effectiveness and superiority of the proposed method. Specifically, the glass panel segmentation achieves an improvement of 0.1927 in mIoU over SOTA methods, and only our method enables instance-level panel extraction. The UAV view planning improves novel view synthesis for reflective façades by 13.15 dB in PSNR compared to commercially used nap-of-the-object planning methods. The RefGlass-GS modeling outperforms SOTA Gaussian Splatting approaches for reflective scenes with an average improvement of 5.08 dB in PSNR. The proposed architecture encompasses the multi-process pipeline from data acquisition and scene modeling to downstream applications, contributing scientific algorithmic innovations while advancing the practical industry adoption of Gaussian Splatting.

**Keywords:** Gaussian Splatting; Reflective Glass Façades; Digital Twin; UAV View Planning; High-Frequency Near-Field Reflection; Glass Panel Segmentation.

## 1 Introduction

Digitizing real-world buildings from captured data is a fundamental task in the development of digital twins and smart cities [1]. A digital twin of a building, defined as its digital replica [2], provides valuable support for facility management across the building lifecycle. Existing building digitization pipelines typically rely on photogrammetry and LiDAR for 3D reconstruction to capture geometric structure and enrich semantic information, and can be further extended to Scan-to-BIM workflows [3] that transform raw observations into object-based, information-rich models. However, for a ubiquitous type of architecture in modern cities, the buildings with reflective glass façades, conventional digitization methods face three major challenges:

(1) **First, geometric reconstruction is unreliable.** The appearance of reflective glass façades is strongly view-dependent, which violates the multi-view consistency assumption required by photogrammetry [4]. As a result, photogrammetric digitization often leads to severe surface geometric distortions and missing structures. LiDAR-based acquisition is also unreliable for glass, as laser beams may penetrate or interact irregularly with the surfaces, resulting in incomplete geometry.

(2) **Second, realistic rendering is inadequate.** Conventional digitization methods typically adopt point clouds or mesh models as the underlying representation [5]. However, both representations are limited to modeling static appearance and are incapable of faithfully representing the high-frequency, view-dependent visual effects of reflective glass façades. Consequently, existing digital models of reflective glass façades often exhibit disordered or unrealistic surface textures.

(3) **Third, semantic modeling is difficult.** Semantic enrichment of digital models for reflective glass façades requires effective object-based segmentation of individual glass panels, which are important components in facility management. However, current state-of-the-art (SOTA) glass segmentation methods [6-9] typically rely on pattern or material differences to segment glass regions from surrounding elements with distinct materials, such as windows embedded in tiled or concrete façades. For an entire façade composed almost exclusively of glass, accurately instance-segmenting each individual panel remains highly challenging.

Recent advances in 3D reconstruction and rendering, particularly the emergence of Neural Radiance Fields (NeRF) [10] and Gaussian Splatting (GS) [11], have provided new opportunities for tackling the digitization challenges of reflective glass façades. In contrast to NeRF, which represents scenes implicitly with multilayer perceptrons (MLPs) and requires computationally intensive ray-based sampling and volume rendering, Gaussian Splatting represents scenes explicitly with learnable Gaussians and renders them through rasterization. This explicit and efficient formulation supports high-fidelity modeling, fast training, and convenient editability [12]. These properties make Gaussian Splatting especially promising for digital twin applications, where both visual realism and interactive usability are essential. However, vanilla Gaussian Splatting models color with spherical harmonics (SH), which are inherently limited to low-frequency view-dependent effects and thus inadequate for representing the high-frequency reflections observed on glass façades. Recent variants [13-16] for reflective scenes have attempted to address this limitation by introducing environment maps to model such high-frequency variations. However, environment maps assume that incident illumination originates from an infinite distance, whereas reflections on real-world reflective glass façades are often dominated by near-field content. As a result, these approaches often suffer from parallax errors.

Meanwhile, for urban building digitization, unmanned aerial vehicles (UAVs) are commonly used for data acquisition. Effective UAV viewpoint planning is critical, as capturing data from appropriate viewpoints directly affects the quality of the final modeling. However, existing UAV viewpoint planning strategies are primarily designed either to maximize the accuracy and completeness of photogrammetric models [17] or to improve the multi-view consistency of Gaussian Splatting reconstruction [18]. For reflective glass façades, however, the imaging requirement is fundamentally different: the captured images should provide comprehensive coverage of the view-dependent appearance of the façade. This UAV viewpoint planning oriented toward view-dependent appearance recovery has not yet been adequately studied. In addition, the final semantically enhanced Gaussian Splatting models are documented as discrete Gaussian primitives rather than the object-based representations typically required in facility management and digital twin platforms [19], which creates the need for effective data organization to support downstream practical applications, such as the interactive management and visualization.

To meet the practical demands of digitizing reflective glass façades for facility management, four key scientific problems must be addressed:

(1) How to segment individual glass panels from an entire glass façade to enable semantic enhancement of GS-based digital models?
(2) How to optimize UAV viewpoint planning to capture observations with comprehensive view-dependent appearance for GS reconstruction?
(3) How to develop an effective and efficient Gaussian Splatting method for modeling high-frequency near-field reflections?
(4) How to organize the resulting GS representation into an object-based model suitable for interactive facility management applications?

To address these challenges, we propose **RefGlass-GS**, an end-to-end fusion framework for the effective and efficient digitization of reflective glass façades, supporting realistic rendering, semantic enhancement, and downstream interactive facility management applications. The main contributions of this work are as follows:

(1) We propose a **novel individual glass panel segmentation method** that formulates façade panel extraction as a maximum a posteriori estimation problem with structural regularities, enabling instance-level segmentation across an entire glass façade under severe reflection and background interference.
(2) We formulate an **effective UAV viewpoint planning optimization function** that maximizes the coverage of view-dependent appearance, thereby improving the suitability of captured data for reflective façade digitization.
(3) We develop an **optimized Gaussian Splatting framework** with a Reflection MLP and two enhanced regularization terms for effective modeling of high-frequency near-field reflections, together with a novel deferred shading function for photorealistic scene rendering.
(4) We introduce a **standardized data mapping paradigm** for organizing GS-based representations into object-based models, bridging the gap between Gaussian primitives and the object-centric data structures required for interactive facility management on practical digital twin platforms.

Experiments on real-world buildings with reflective glass façades demonstrate the effectiveness of our RefGlass-GS. In glass panel segmentation, our method outperforms SOTA deep-learning-based glass segmentation approaches with an mIoU improvement of 0.1927 and enables reliable instance-level panel extraction. In UAV viewpoint planning, the proposed data acquisition strategy captures view-dependent reflective appearance more comprehensively than commercially used nap-of-the-object view planning methods, thereby improving novel view synthesis for reflective façades by 13.15 dB in PSNR. In Gaussian Splatting-based modeling, our method achieves higher-fidelity and more photorealistic results than SOTA GS approaches for reflective scenes, with an average improvement of 5.08 dB in PSNR. To the best of our knowledge, RefGlass-GS is the first end-to-end fusion framework to unify multiple processes, ranging from data acquisition and model reconstruction to downstream application, within a Gaussian Splatting pipeline for real-world reflective glass façade scenarios and digital twin development. The proposed architecture contributes not only novel algorithms, but also a practical pathway for bringing Gaussian Splatting into real-world industrial applications.

## 2 Literature Review

This section provides a comprehensive review of prior research in the three areas closely relevant to our contributions and identifies the limitations of existing methods. It is organized as follows: (2.1) Gaussian Splatting and Variants for Reflective Scenes; (2.2) Glass Segmentation; and (2.3) UAV View Planning.

### 2.1 Gaussian Splatting and Variants for Reflective Scenes

As a breakthrough in neural reconstruction and rendering after the advent of NeRF, Gaussian Splatting has received substantial attention and has developed rapidly in recent years. This research direction began in 2023 with the introduction of 3D Gaussian Splatting (3DGS) by Kerbl et al. [11]. The success of 3DGS can be attributed to several key design choices, including (1) the use of 3D Gaussian ellipsoids to represent scene geometry and appearance, (2) an optimization strategy that jointly learns Gaussian attributes while dynamically controlling the number of primitives, and (3) a GPU-accelerated rasterization pipeline that exploits parallel computation to achieve fast and efficient rendering. Benefiting from these properties, 3DGS supports photorealistic reconstruction, real-time rendering, efficient training, and flexible editing. These advantages make Gaussian Splatting highly promising for the construction of digital twin models.

Nevertheless, the volumetric representation of 3D Gaussian ellipsoids in 3DGS may introduce multi-view inconsistencies in estimated surface normals, thereby limiting the reconstruction of accurate geometric surfaces. To alleviate this issue, 2D Gaussian Splatting (2DGS) [20] replaces volumetric Gaussian primitives with Gaussian elliptical disks that approximate mesh-like triangles, leading to improved multi-view consistency and more accurate surface geometry. Besides, recent studies have also explored semantic enhancement for GS-based models. Gaussian Grouping [21] first segments the input images using the Segment Anything Model (SAM) [22], and then assigns unique object identities to Gaussian primitives during training via mask association and 3D spatial consistency regularization. Subsequently, several language-embedded GS methods [23,24] have incorporated zero-shot foundation models to extract high-level semantic features, such as CLIP [25] or DINO [26] features, and embed them into Gaussian representations, enabling natural-language interaction with GS models. These advances in geometric accuracy and semantic enrichment further support the use of Gaussian Splatting as a representation for digital twin models.

However, current GS-based methods still face significant challenges when applied to more complex scenes, especially those containing high-frequency reflective surfaces. Conventional GS-based methods without dedicated reflection modeling, such as 3DGS [11] and 2DGS [20], typically rely on spherical harmonics for color representation, which are only suitable for modeling low-frequency view-dependent effects. To better capture high-frequency reflective appearance, several studies have extended Gaussian Splatting with environment maps. One of the earliest works, GaussianShader [13], estimates surface normals from the shortest-axis directions of 3D Gaussians and employs a simplified shading function to query an environment map for reflection rendering. 3DGS-DR [15] improves this strategy by introducing normal propagation during training, stabilizing normal-related gradients and improving the reconstructed geometry used for environment-map queries. Ref-Gaussian [14] instead builds upon 2DGS, taking advantage of its relatively stable surface reconstruction to estimate normals and query the environment map, while further adopting a physically based deferred rendering scheme for reflection synthesis. MaterialRefGS [27] introduces a multi-view material consistency constraint and a reflection strength prior to generate multi-view consistent material maps during deferred shading, thereby facilitating more accurate material inference for environment-map-based shading. Despite these advances, environment-map-based reflection modeling assumes that global illumination originates from infinity, which is generally only suitable for synthetic scenes. In real-world scenes, however, reflections are often dominated by near-field objects and lighting conditions, leading to parallax errors in these environment-map-based methods. Subsequent studies attempted to address the limitation of near-field illumination through ray tracing. For example, SpecTRe-GS [28] first reconstructs the object and scene using 3DGS and an environment map, and then incorporates secondary rays to model reflections caused by nearby objects. IRGS [16] further proposes a differentiable 2D Gaussian ray tracer to compute incident radiance on the fly, and combines it with the full rendering equation to synthesize high-quality reflections from nearby objects. Nevertheless, these methods [16,28] primarily target inter-reflections between objects. In particular, they require explicit reconstruction of the surrounding objects responsible for near-field reflections, after which ray tracing is used to query their reflected appearance on the target reflective surface. Therefore, they are unable to recover near-field reflections solely from observations of the reflective surface itself. This makes them more suitable for small-scale object-centric scenes, but less practical for large real-world scenes, where reconstructing all reflection-related surroundings is computationally expensive. Additionally, Ref-GS [29] incorporates surface light fields by enriching directional encoding with spatial information and estimating the reflection through a compact MLP; however, it still struggles to faithfully reproduce high-frequency details.

Therefore, current GS-based modeling methods remain unable to accurately digitize the high-frequency near-field reflection effects of reflective glass façades. To address this gap, our proposed RefGlass-GS introduces a Reflection MLP combined with a deferred shading function to better model complex reflections beyond previous GS-based approaches. By further integrating semantic enhancement, RefGlass-GS provides a more effective framework for the digitization of reflective glass façades.

## 2.2 Glass Segmentation

Glass segmentation has long been recognized as a challenging problem in computer vision due to the unique optical properties of glass. Unlike diffuse materials, glass often lacks stable intrinsic texture while simultaneously exhibiting appearance patterns induced by the surrounding environment. Existing methods have increasingly adopted deep-learning paradigms to exploit learned contextual cues for this task. Representative works include GDNet [30], which constructs a large-scale Glass Detection Dataset (GDD) and learns abundant multi-scale contextual features to distinguish glass regions in real-world scenes. Building upon this, EBLNet [6] emphasizes enhanced boundary learning through refined edge supervision to improve the segmentation of glass-like objects such as windows and mirrors. GhostingNet [7] further leverages intrinsic ghosting effects arising from double reflections within glass layers via a dedicated Ghosting Effects Detection module to guide surface segmentation. Similarly, GlassSemNet [8] incorporates semantic relations between glass and surrounding objects to resolve ambiguities in complex scenes, while RFENet [31] introduces reciprocal feature evolution between boundary and contextual representations for robust segmentation of glass-like objects. Complementary approaches further enhance glass segmentation by aggregating multi-level contextual information and reflection cues [32] or by employing edge-guided lightweight architectures to improve boundary precision [33]. However, these methods can achieve good performance by modeling the intrinsic differences between glass and surrounding materials, such as low texture variance, reflection inconsistencies, and contextual discontinuities. And they typically formulate the task as semantic segmentation, producing binary or multi-class masks that classify pixels as glass or non-glass. For reflective glass façades, however, where most regions belong to glass and material distinctions between adjacent panels are often weak or even negligible, such methods tend to segment the entire curtain wall as a single glass region, rather than performing instance-level segmentation to differentiate individual glass panels.

More recently, specialized deep-learning methods have been developed for curtain wall frame segmentation to localize panel boundaries more directly. For example, LANet [34] proposes a lightweight attention network with an optimized transformer module for real-time curtain wall frame segmentation, designed to support automated façade installation by robots or crane operators in high-rise construction. Similarly, C2PNet [35] introduces a context collaboration pyramid network that combines a pyramid pooling transformer encoder with context interaction and channel-guided pyramid convolution decoders, enabling accurate frame segmentation from single RGB images for curtain wall installation robots. However, these methods still rely heavily on the pronounced visual and material differences between frames and glass to achieve accurate frame segmentation. As evidenced by the datasets prepared in these studies, such approaches generally assume frames with clearly visible width and distinct appearance, which may be suitable for windows or glass doors but are less applicable to reflective glass façades, where frame structures are often intentionally minimized or visually inconspicuous for aesthetic design. Moreover, the datasets used in both works are mainly composed of planar glass curtain walls, limiting their applicability to curved curtain wall structures.

Therefore, current glass segmentation methods remain inadequate for instance-level segmentation of reflective glass façades when only subtle material differences exist between adjacent panels. Such a limitation is detrimental to object-based semantic enhancement for façade digitization, where accurate panel-wise understanding is essential. To address this gap, our proposed framework formulates façade panel segmentation as a maximum a posteriori estimation problem using structural regularities, thereby enabling robust instance-level segmentation across the entire glass façade under severe reflection and background interference.

### 2.3 UAV View Planning

UAV view planning plays a critical role in autonomous data acquisition for high-quality 3D reconstruction [36,37]. Most existing approaches have been developed for photogrammetric reconstruction, with the goal of achieving accurate and complete 3D models through optimized viewpoint selection and trajectory design. For example, Liang et al. [38] proposed an optimized UAV view planning framework based on a modified sparrow search algorithm, which integrates exploration-exploitation stages with semantic and geometric analysis to generate minimal yet high-quality viewpoint sets for building reconstruction. Peng et al. [39] presented an adaptive view planning algorithm that starts from coarse proxy models and iteratively selects viewpoints on a view manifold, balancing coverage and sparsity for aerial 3D reconstruction. Similarly, Shang et al. [40] introduced a topology-based UAV path planning method for multi-view stereo reconstruction of complex structures, decomposing the problem into parallelizable overlapped-view optimization subproblems that exploit structural topology to reduce computational cost while improving reconstruction fidelity. Zheng et al. [41] developed a cooperative route planning method for multiple UAVs, computing minimum image sets and optimal flight networks under battery and time constraints to obtain fine-resolution building models. Hepp et al. [42] introduced Plan3D, a viewpoint and trajectory optimization framework for aerial multi-view stereo that efficiently computes globally consistent paths in outdoor environments. Model-quality-guided planning was further explored by Zhang et al. [43], who designed a UAV path planning strategy driven by real-time reconstruction quality metrics to ensure complete and accurate 3D models of complex buildings. Specialized applications include the automated UAV path planning method proposed by Wang et al. [44] for high-quality photogrammetric 3D bridge reconstruction and the flight path planning strategy developed by Liu et al. [45] for UAV-based refinement inspection of construction sites, both of which iteratively refine trajectories using reconstruction feedback.

However, view planning for Gaussian-based reconstruction remains at an early stage. Unlike traditional photogrammetry, Gaussian Splatting demands viewpoints that not only capture geometry but also optimize the density and covariance of Gaussian primitives for high-fidelity rendering. Recent works have begun to extend active reconstruction principles to this setting. For instance, GS-Planner [18] proposes a Gaussian-Splatting-based planning framework for active high-fidelity reconstruction that performs uncertainty estimation and next-best-view selection directly on the evolving 3DGS representation. Building on hierarchical decomposition, HGS-Planner [46] evaluates reconstruction completion and quality gains via Fisher information and unobserved regions in the Gaussian representation, and integrates global and local planners for efficient autonomous exploration. ActiveGS [47] employs a hybrid Gaussian–voxel map for active scene reconstruction using RGB-D inputs on mobile platforms to enable accurate mapping of unknown environments. ActiveSplat [48] proposes an autonomous high-fidelity reconstruction system that combines efficient rendering with Voronoi-based exploration graphs within the Gaussian Splatting pipeline. Furthermore, Wu et al. [49] investigated optimal drone trajectory planning for capturing 3DGS objects by formulating an online utility-based optimization problem to maximize synthesized view quality under time constraints. Nevertheless, existing Gaussian Splatting view planning methods are primarily designed for ordinary diffuse scenes with relatively view-independent appearance. They cannot be directly applied to reflective glass façades, where appearance is strongly view-dependent due to specular reflections. For high-fidelity reconstruction and photorealistic rendering of glass reflections, captured images must provide sufficiently diverse angular observations of the reflected content; otherwise, limited viewpoint coverage may result in incomplete or inaccurate reflection modeling in the Gaussian representation [50].

Therefore, current UAV view planning methods remain insufficient for acquiring data suitable for high-fidelity reconstruction and rendering of reflective glass façades, whose appearance varies significantly with viewpoint. To address this gap, our proposed framework constructs a unit hemisphere model over each glass panel and formulates a novel optimization objective that maximizes the coverage of view-dependent appearance, thereby enabling more complete observation of reflection behavior during data acquisition.

## 3 Methodology

The overall architecture of our proposed RefGlass-GS fusion framework is illustrated in Fig. 1. The entire framework consists of the following five steps:

(1) In Step 1, we first conduct an initial UAV flight using automated simple flight planning functions, such as orbit flight, to reconstruct coarse models of the target building with reflective glass façades, including a coarse point cloud and a coarse mesh model. These coarse models provide a fundamental understanding of the site for subsequent steps.

(2) In Step 2, we perform corner point extraction on a fronto-parallel view image captured during Step 1, based on maximum a posteriori estimation with structural regularities. The extracted corner points are then projected onto the 3D coarse mesh model to obtain 3D glass panel models. For non-planar curved façades, we further propose a rectification method to make the glass panel segmentation robust to curved surface scenarios.

(3) In Step 3, building upon the 3D glass panel models obtained in Step 2, we formulate a UAV view planning optimization problem that maximizes the coverage of view-dependent texture. Specifically, a unit hemisphere model is constructed on each panel to transform the physical coverage objective into a mathematical formulation, based on which a second UAV flight is planned and executed for data acquisition.

(4) In Step 4, we utilize the second-flight images captured in Step 3 to perform RefGlass-GS modeling. By constructing augmented 2D Gaussian primitives together with a Reflection MLP, a novel deferred shading function, and two regularization terms, our method achieves the modeling of high-frequency near-field reflections and the embedding of panel-level semantics, thereby enabling photorealistic and semantic digitization of reflective glass façades.

(5) In Step 5, we organize the GS digital models generated in Step 4 through the proposed standardized object-based data organization scheme, enabling compliance with facility management requirements. The organized models are then imported into a practical interactive digital twin platform, ultimately achieving the full pipeline from reality capturing to digital applications.

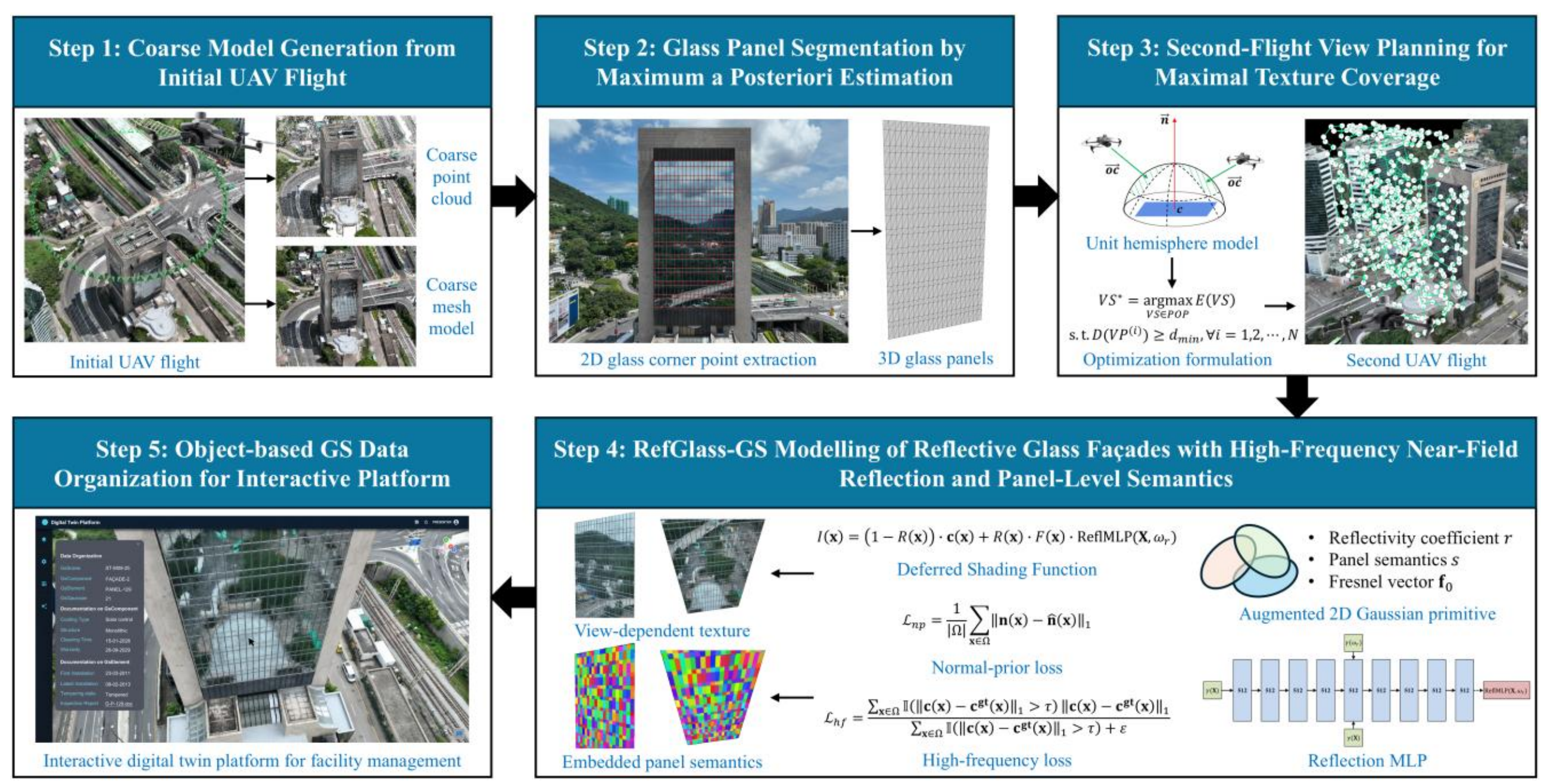

Fig. 1. Overall architecture of the proposed RefGlass-GS fusion framework

### 3.1 Glass Panel Segmentation based on Structural Regularities

To achieve a generalizable glass panel segmentation approach applicable to both planar and curved glass façades, we first describe in Section 3.1.1 how curved glass façades can be effectively rectified into planar surfaces. Subsequently, Section 3.1.2 presents a Maximum A Posteriori (MAP) estimation framework that exploits structural regularities to accomplish individual glass panel segmentation.

### 3.1.1 Image Rectification for Curved Surfaces

For curved façades, an image rectification step is first applied to warp the curved surface into a planar representation in the image domain, thereby facilitating subsequent segmentation. Fig. 2 illustrates the image rectification process for curved façades. A close-range frontal-view image is captured near the façade surface, containing only the façade region without background clutter. By exploiting the 3D-to-2D projection relationship between the coarse façade point cloud (with spatial coordinates $(X, Y, Z)$) and the corresponding image pixels $(u, v)$, as defined in Eq. (1), those 3D points whose projections fall within the image extent are identified and isolated to construct a cropped point cloud. A cylindrical surface model is subsequently fitted to this cropped point cloud through least-squares optimization [51], producing the cylinder parameters that describe the curved façade geometry.

$$\begin{bmatrix} u \\ v \\ 1 \end{bmatrix} = \mathbf{K} \cdot [\mathbf{R}|\mathbf{T}] \cdot \begin{bmatrix} X \\ Y \\ Z \\ 1 \end{bmatrix} \tag{1}$$

where **K** is the camera intrinsic matrix; **R** and **T** denote the extrinsic rotation matrix and translation vector, respectively.

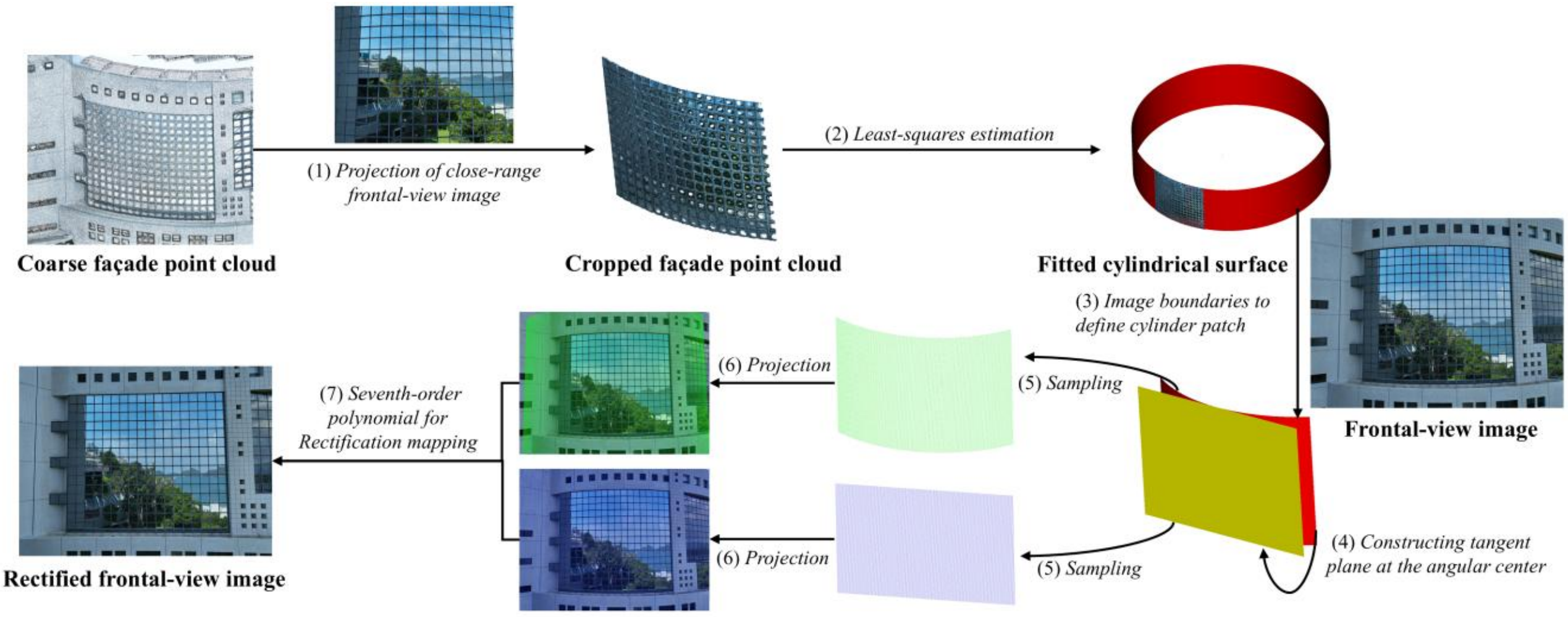

Fig. 2. Image rectification pipeline for curved façade surfaces

For the image on which panel segmentation is to be performed, the image boundaries serve to identify the corresponding region on the fitted cylindrical surface, thus delineating a cylindrical patch (indicated in red in the figure). This patch is then discretized to generate a set of surface sample points (green points). A tangent plane is subsequently established at the angular midpoint of the patch (yellow plane), onto which the sampled surface points are orthogonally projected, yielding a flattened point set (blue points).

Both the original surface samples on the cylinder (green points) and their corresponding planar projections (blue points) are then mapped onto the target image. Within this mapping, the green points represent pixel positions associated with the actual 3D curved façade geometry, while the blue points represent the pixel positions that result from unwarping the curved surface into a flat configuration. Hence, transforming a curved-façade image into an equivalent planar-façade image reduces to estimating a coordinate mapping from the green-point pixel positions to the blue-point pixel positions. A seventh-order bivariate polynomial is adopted to model this transformation, achieving robust image rectification for curved façades. The transformation is formulated as:

$$u' = \sum_{k=0}^{7} \sum_{l=0}^{7-k} a_{kl} \, u^k v^l$$
$$v' = \sum_{k=0}^{7} \sum_{l=0}^{7-k} b_{kl} \, u^k v^l \qquad (2)$$

where $(u, v)$ are the original pixel coordinates on the curved-façade image (green points), $(u', v')$ are the corresponding rectified pixel coordinates on the planar-façade image (blue points), and $a_{kl}$, $b_{kl}$ are polynomial coefficients estimated via least-squares fitting from the paired point correspondences. The total number of coefficients for each coordinate is $\binom{7+2}{2} = 36$, resulting in 72 parameters in total.

### 3.1.2 Maximum a Posteriori Estimation for Individual Glass Panel Segmentation

After transforming all images of reflective glass façades into rectified fronto-parallel views, where grid lines are maximally aligned with the image axes, we formulate the panel segmentation task as a Maximum A Posteriori (MAP) estimation problem that exploits the structural regularities of glass panels on the façade, specifically leveraging the structural linear features delineating panel boundaries and the repetitive patterns formed by these lines.

To characterize the structural linear features, we construct structure-line intensity histograms along both the height and width directions of the image. As illustrated in Fig. 3, discrete line features are first extracted from the image using the Probabilistic Hough Transform [52]. These line segments are then classified into horizontal or vertical orientations by estimating their respective vanishing points through a RANSAC-based procedure [53]. The structure-line intensity histograms record, for each row or column of the image, the cumulative count of pixels that belong to detected line features. Formally, we define two vectors: $HLI$ and $VLI$. The dimension of $HLI$ corresponds to the image height $H$ and encodes the horizontal line intensity, whereas the dimension of $VLI$ corresponds to the image width $W$ and encodes the vertical line intensity. Specifically, $HLI(i)$ denotes the cumulative pixel length of all horizontal line features located at row $i$, and $VLI(j)$ denotes the cumulative pixel length of all vertical line features located at column $j$. As shown in Fig. 3, the orange bars along the vertical axis represent $HLI$, while the blue bars along the horizontal axis represent $VLI$.

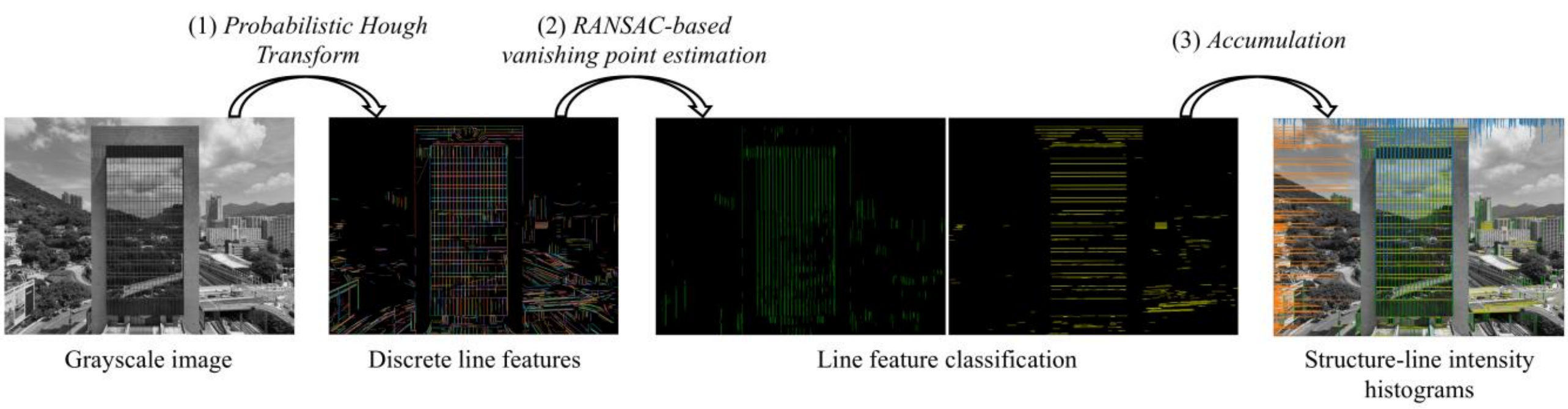

Fig. 3. Structural linear feature extraction

Although each peak in the structure-line intensity histograms can be interpreted as a potential boundary between adjacent glass panels, the significant interference caused by reflective textures and background noise in reflective glass façade scenarios can lead to substantial errors when directly segmenting panels based on structural line intensities alone. Instead, we regard these peaks as prior probabilities indicating regions where boundaries are most likely to occur. Furthermore, by incorporating the repetitive patterns inherent in the structural line arrangements, we formulate a MAP estimation framework that enables more robust segmentation of individual glass panels

We define two similarity matrices, $HSIM$ and $VSIM$, to capture the repetitive patterns of structural lines along the horizontal and vertical directions, respectively. The dimension of $HSIM$ is $H \times H$, and that of $VSIM$ is $W \times W$. Taking $HSIM$ as an example, let $C_i$ denote all pixels at row $i$ of the image (i.e., the $i$-th row), and $C_j$ denote all pixels at row $j$ (i.e., the $j$-th row). Then, $HSIM(i,j)$ quantifies the similarity between the $i$-th and $j$-th rows, computed using the Structural Similarity Index (SSIM) [54] as defined in Eqs. (3) and (4). An illustration of $HSIM$ is provided in Fig. 4.

$$HSIM\ (i,j) = SSIM(C_i, C_j) = \frac{(2\mu_i\mu_j + c_1)(2\sigma_{ij} + c_2)}{({\mu_i}^2 + {\mu_j}^2 + c_1)({\sigma_i}^2 + {\sigma_j}^2 + c_2)}, \text{where } 0 < i,j \leq H \text{ and } i,j \in \mathbb{Z} \quad (3)$$

$$VSIM\ (i,j) = SSIM(C_i, C_j) = \frac{(2\mu_i\mu_j + c_1)(2\sigma_{ij} + c_2)}{({\mu_i}^2 + {\mu_j}^2 + c_1)({\sigma_i}^2 + {\sigma_j}^2 + c_2)}, \text{where } 0 < i,j \leq W \text{ and } i,j \in \mathbb{Z} \quad (4)$$

where, $\mu_i$ and $\mu_j$ denote the mean values of $C_i$ and $C_j$ respectively, while $\sigma_i$ and $\sigma_j$ represent their standard deviations. $\sigma_{ij}$ indicates the covariance between $C_i$ and $C_j$. Constants $c_1$ and $c_2$ are introduced to avoid numerical instability caused by a denominator of zero.

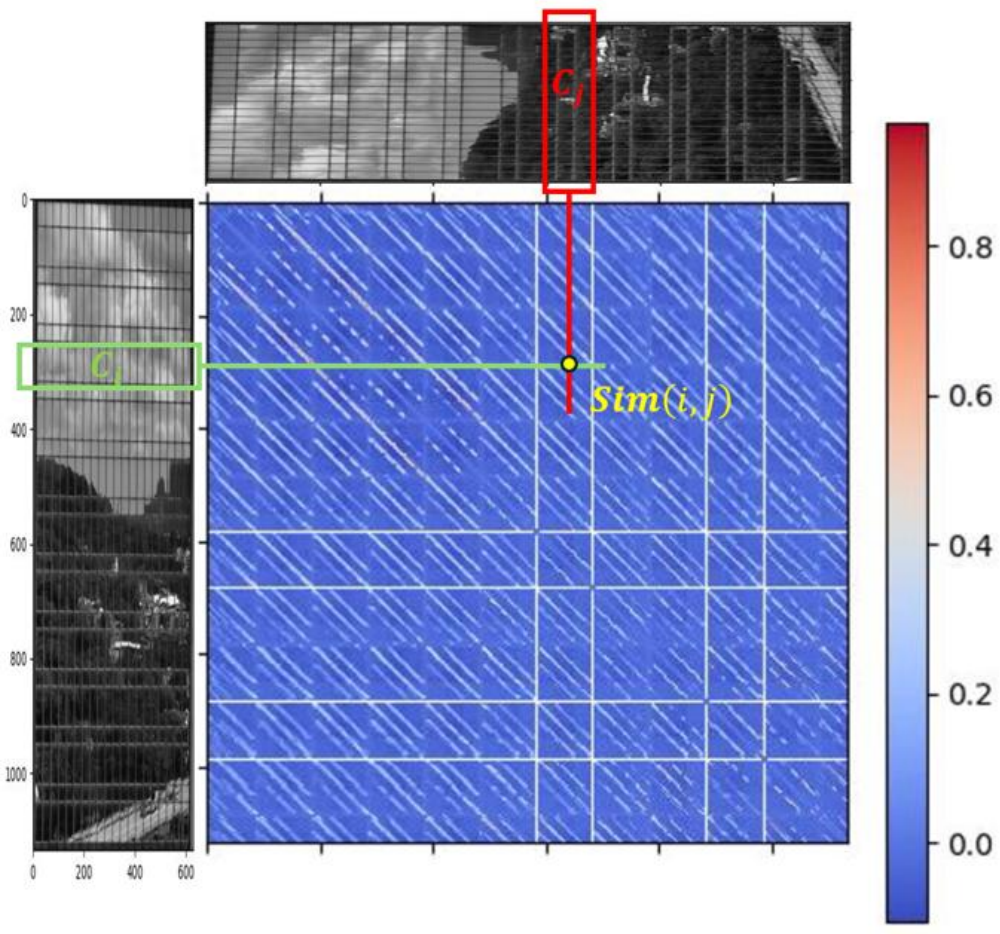

Fig. 4. Repetitive patterns represented by similarity matrices

After that, the structural line intensities are used as the prior probability, and the similarity matrices are used as the likelihood function to construct the MAP models $HPosterior$ and $VPosterior$ for identifying glass panel corner point at horizontal and vertical directions, as shown in Eqs. (5) and (6).

$$HPosterior(i) = \sum_{j=1}^{H} NorHSIM(i,j) \times NorHLI\ (j) \quad (5)$$

$$VPosterior(i) = \sum_{j=1}^{W} NorVSIM(i,j) \times NorVLI(j) \quad (6)$$

where $NorHSIM$ and $NorVSIM$ denote the normalized similarity matrices, while $NorHLI$ and $NorVLI$ denote the normalized structure-line intensities histograms. Taking the horizontal direction as an example, the normalization procedures are defined as follows:

$$NorHLI\ (j) = \frac{HLI(j)}{\max_{k \in H} HLI(k)} \quad (7)$$

$$NorHSIM(i,j) = \frac{\exp\left(\beta \cdot HSIM(i,j)\right)}{\sum_{j \in H} \exp\left(\beta \cdot HSIM(i,j)\right)} \quad (8)$$

Here, $\beta$ is a temperature parameter that controls the sharpness of the softmax normalization, and is set to 20 in our experiments.

The local maxima identified in $HPosterior$ and $VPosterior$ correspond to the corner points of the glass panels, as illustrated in Fig. 5(a). For curved façades, the corner points extracted in the rectified image are further mapped back to the unrectified image using the fitted seventh-order polynomial as shown in Eq. (2).

Then, we project the glass corner points from the 2D image domain onto the initial 3D coarse mesh model via ray casting. Because corner extraction in the image may incur subtle pixel-level errors, which can be amplified after ray casting over distances of tens of meters, the resulting 3D panel corners are regularized: for each row of corners, the $z$-coordinates are averaged, and for each column, the $x$- and $y$-coordinates are averaged, thereby reducing segmentation-induced deviations. The 3D panel models are then obtained by pairing the regularized corner points, as shown in Fig. 5(b).

The proposed glass panel segmentation approach demonstrates robustness against reflections and distortions. Moreover, the method accommodates scenarios where glass panel sizes vary, and panels arranged along each direction are not required to share identical dimensions. This flexibility arises because the extraction of repetitive patterns is agnostic to the specific spacing between line features, thereby generalizing to façades with non-uniform panel configurations.

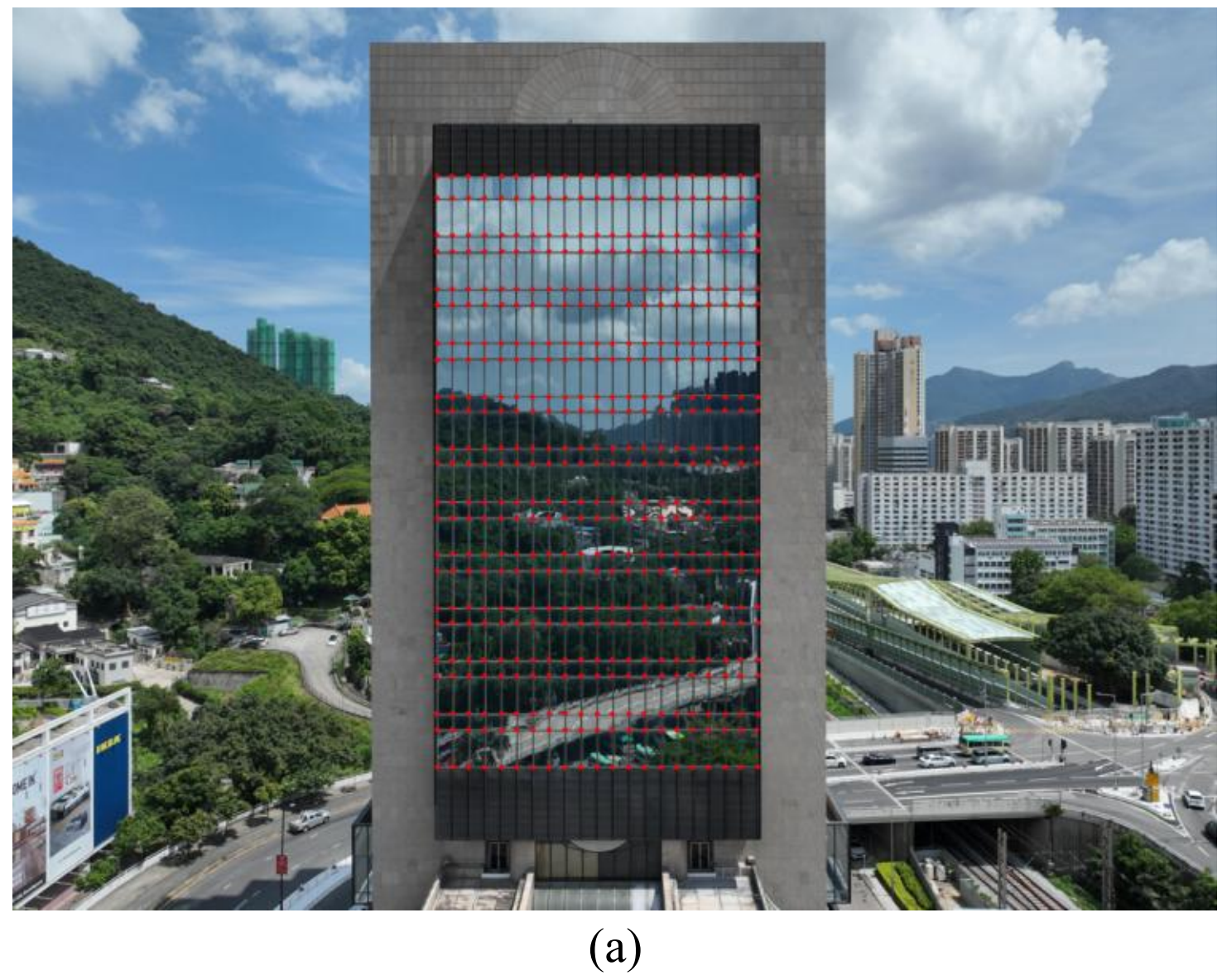

(a)

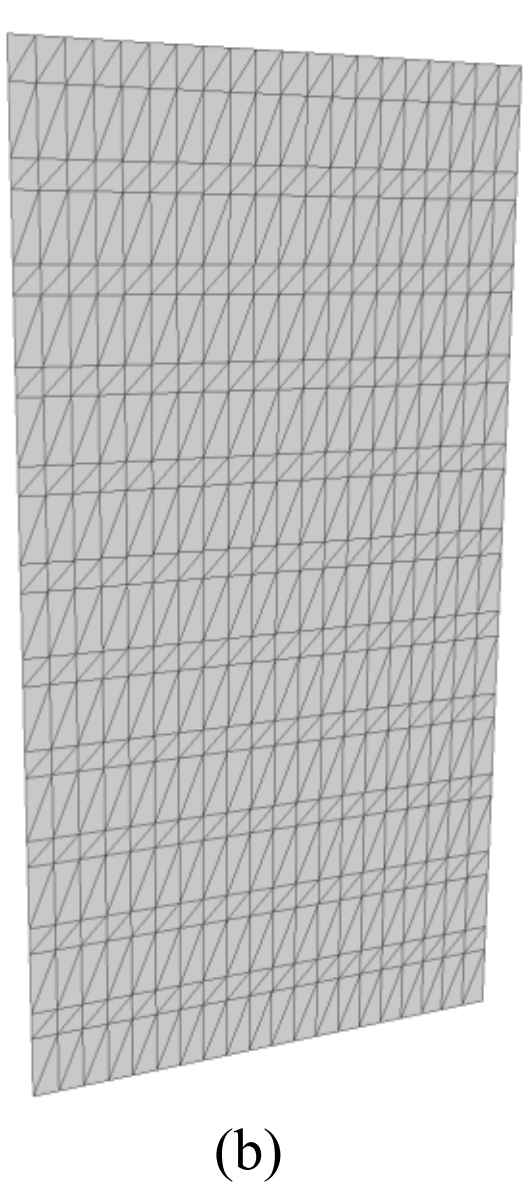

(b)

Fig. 5. Glass panel segmentation results: (a) 2D corner point extraction by MAP estimation; (b) 3D panel model generation through ray casting

### 3.2 UAV View Planning Formulation for Maximal Texture Coverage

The objective of our UAV view planning is to maximize the coverage of view-dependent reflective textures on the glass façade, thereby enabling the recovered model to render more photorealistic scenes. This overarching goal can be discretized equivalently as ensuring that the view-dependent texture of each individual glass panel on the façade is observed and captured from as many perspectives as possible.

Therefore, we formulate the objective of maximal view-dependent texture coverage on each individual 3D glass panel. For each panel, we first extract its centroid $c$ and surface normal vector $\vec{n}$. A unit hemisphere oriented along $\vec{n}$ is then constructed with $c$ as the origin. Achieving maximal texture coverage for each glass panel can thus be equivalently expressed as ensuring that the UAV's lines of sight collectively cover the unit hemisphere of each panel as completely as possible.

To this end, we establish a spherical coordinate system centered at the panel centroid $c$, and partition the unit hemisphere according to the polar angle $\theta$and azimuthal angle $\varphi$. The hemisphere is discretized at 30-degree intervals for both $\theta$ and $\varphi$, yielding 36 sections per panel hemisphere. Let $o$ denote the optical center of the UAV camera. The line connecting $o$ to $c$ defines a viewing direction $\overrightarrow{oc}$. A section of the panel's hemisphere is considered covered if $\overrightarrow{oc}$ passes through that section and $\overrightarrow{oc}$ lies within the field of view of the UAV camera, as illustrated in Fig. 6.

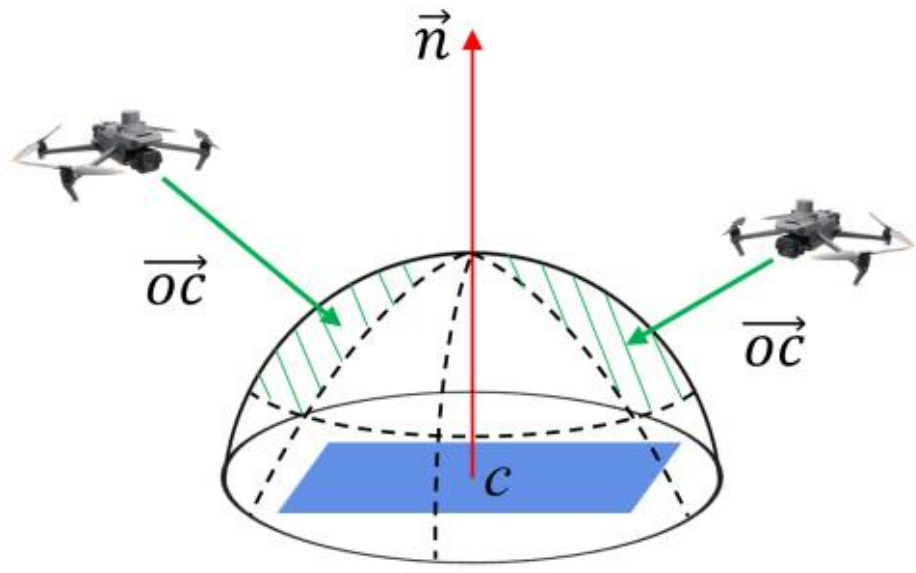

Fig. 6. Unit hemisphere model for UAV view planning formulation

Accordingly, we formulate this as an optimization problem. Since UAV view planning constitutes an Orienteering Problem with NP-hard computational complexity [55], we employ a metaheuristic swarm-based optimization algorithm to solve it. The optimization population is defined as $POP$, where each individual in the swarm represents a viewpoint set $VS$. Each viewpoint set contains $N$ viewpoints $VP$, and each $VP$ is parameterized by five degrees of freedom: position and orientation $(X, Y, Z, \text{pitch}, \text{yaw})$. The optimization aims to identify the optimal viewpoint set $VS^*$ that maximizes the overall texture coverage, as formulated in Eq. (9).

$$\begin{gathered} VS^* = \underset{VS \in POP}{\operatorname{argmax}} E(VS) \\ \text{s.t.}\, D(VP^{(i)}) \geq d_{min}, \forall i = 1,2,\cdots,N \end{gathered} \tag{9}$$

where, $D(VP^{(i)})$ denotes the minimum distance from the $i$-th viewpoint in the viewpoint set to the façade, and $d_{\text{min}}$ represents the safety distance threshold, which is set to 10 meters.

Let $HS^{(jk)}$ denote the $k$-th hemisphere section on the $j$-th glass panel. The fitness function $E$ that evaluates the texture coverage of a viewpoint set is defined in Eq. (10):

$$E(VS) = \sum_{j=1}^{M} \sum_{k=1}^{36} \left[\!\left[ \sum_{i=1}^{N} \text{Interaction}(VP^{(i)}, HS^{(jk)}) \right]\!\right] \tag{10}$$

where $M$ is the number of glass panels; $\text{Interaction}(VP^{(i)}, HS^{(jk)})$ evaluates whether the viewing direction $\overrightarrow{oc}$ of viewpoint $VP^{(i)}$ intersects with hemisphere section $HS^{(jk)}$. To avoid the computational burden of ray-mesh intersection tests, we instead project $\overrightarrow{oc}$ into the panel's spherical coordinate system to obtain its angular coordinates $\theta^{ij}$ and $\varphi^{ij}$, and determine whether they fall within the $\theta$ and $\varphi$ bounds of the corresponding hemisphere section. If so, the viewpoint is considered to intersect with that section. The operator $[\![\cdot]\!]$ is an indicator function that returns 1 if at least one viewpoint in the set intersects with a given hemisphere section, and 0 otherwise. In this manner, the fitness function $E$ counts the total number of hemisphere sections covered across all panels.

The particle swarm optimization (PSO) algorithm [56] is employed to solve this optimization formulation. Viewpoint set candidates are initialized as the population with a population size of $POP = 50$, where each viewpoint is constrained to lie beyond the minimum safety distance $d_{\text{min}}$ from the façade. The number of viewpoints $N$ within each viewpoint set is determined through the parametric analysis presented in Section 4.2.1. The optimization process is executed for a total of 200 iterations; experimental results demonstrate that this number of iterations is sufficient for the optimization to converge.

The optimized viewpoint set $VS^*$ obtained through the proposed UAV view planning is subsequently used to solve a Traveling Salesman Problem (TSP) via a greedy algorithm [38], generating the shortest obstacle-avoidance flight trajectory that traverses all viewpoints. The resulting viewpoints and trajectory are then loaded into the UAV to conduct a second flight for image acquisition.

### 3.3 RefGlass-GS Modeling of Reflective Glass Façades

The image dataset acquired from the second flight is further utilized to perform photorealistic and semantic digitization via Gaussian Splatting, while also supporting subsequent interactive functionalities. The captured images first undergo the preprocessing pipeline described in Section 3.3.1, followed by the modeling procedures detailed in Sections 3.3.2 through 3.3.4.

#### 3.3.1 Preprocessing of the Second Flight Captured Data

We first perform glass panel semantic segmentation on the images captured during the second flight. Unlike repeating the glass panel segmentation procedure of Section 3.1 for each individual image, we leverage the 3D panel models obtained in Section 3.1.2, combined with the geo-coordinates and orientation of each photograph, to conduct 3D semantic back-projection via Eq. (1). The resulting back-projected segmentation maps are illustrated in Fig. 7. Notably, the semantic label assigned to each glass panel is stored solely as its panel index, that is, each distinct color in the segmentation map corresponds to a unique glass panel index. This compact storage scheme minimizes memory consumption and simultaneously facilitates the panel numbering required for the interactive facility management presented in Section 3.4.

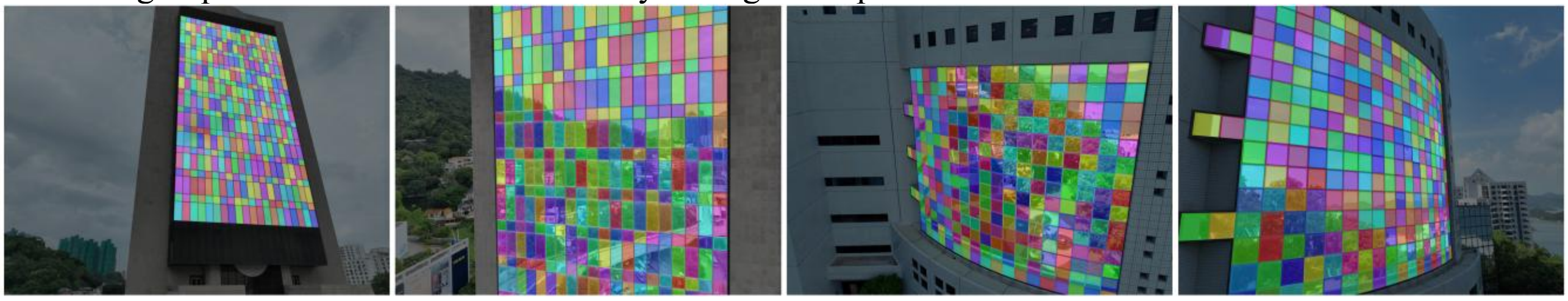

Fig. 7. Semantic back-projection for glass panel segmentation on second-flight images

Subsequently, we aim to obtain sufficiently accurate normal maps to serve as the basis for the Reflection MLP estimation described in Section 3.3.3. The captured images are processed through oblique photogrammetric reconstruction to generate a corresponding 3D mesh model. However, this mesh model exhibits geometric distortions in the reflective glass façade regions. Simply overlaying the 3D panel models obtained in Section 3.1.2 onto this mesh model still results in an uneven façade surface (Fig. 8(a)), as the mesh often contains local protrusions and irregularities within the panel regions. On the other hand, naively substituting the corresponding mesh regions with the 3D panel models would inadvertently remove the surrounding frame structures. To overcome this problem, each 3D panel undergoes a slight in-plane contraction toward its center, and a small cuboid-shaped region is constructed along the normal of the panel. The region of the coarse façade mesh falling within this cuboid is then replaced by the contracted 3D panel geometry. This operation produces the refined façade surface depicted in Fig. 9, thereby achieving geometrically faithful reconstruction of reflective glass façades (Fig. 8(b)). The refined façade model is subsequently rendered at each camera pose to produce a corresponding normal map. In addition, a façade region mask is rendered to delineate the façade extent, enabling the complex view-dependent reflective regions to be distinguished from the diffuse background areas for separate modeling and rendering.

The complete set of data prepared for subsequent digitization comprises: RGB images, panel semantic segmentation maps, refined normal maps, façade mask images, camera poses, and the point cloud model.

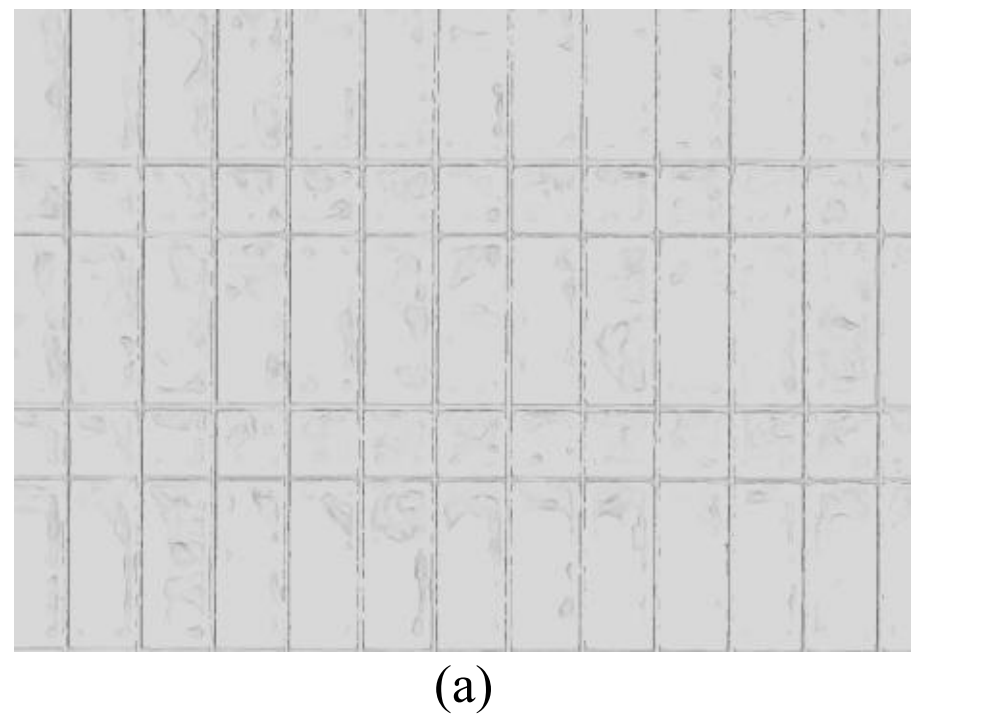

(a)

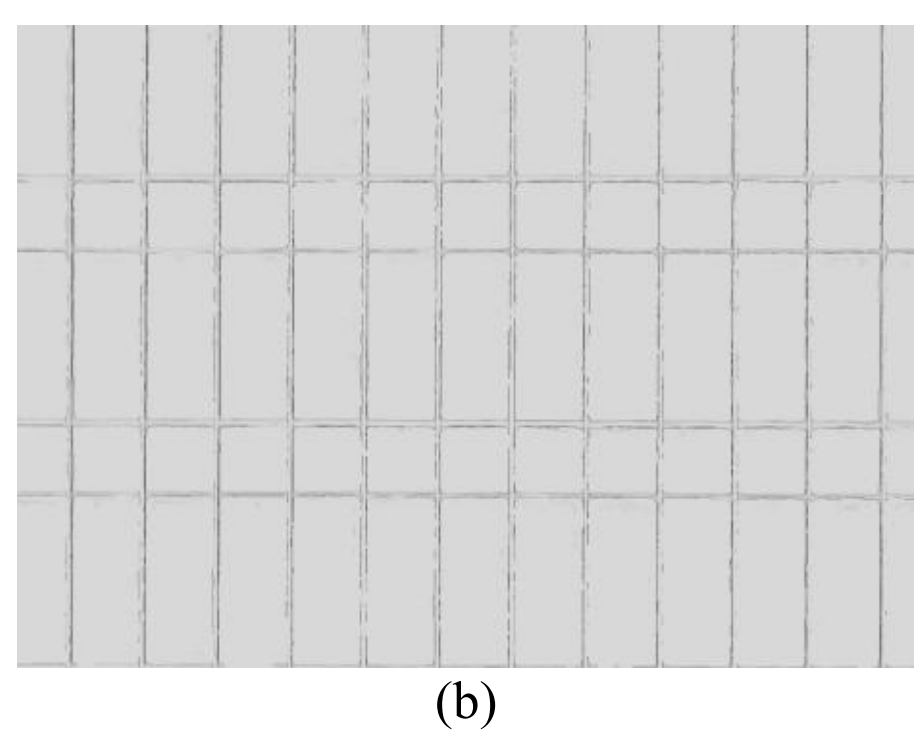

(b)

Fig. 8. Refined mesh by (a) simply overlapping and (b) cuboidal region replacement

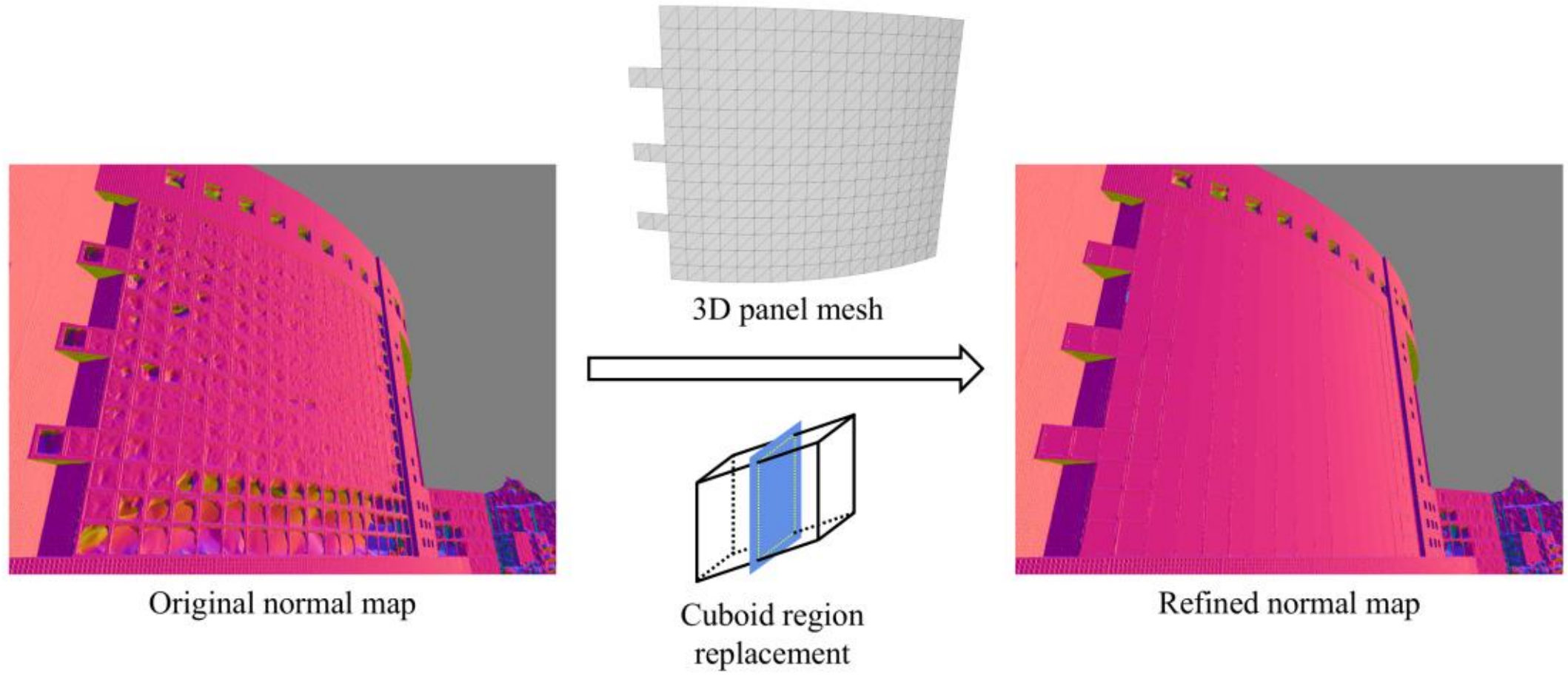

Fig. 9. Cuboidal region replacement for façade surface refinement

### 3.3.2 Augmented 2D Gaussian Splatting and Deferred Shading Function

We employ 2D Gaussian Splatting (2DGS) [20] as the foundational primitive for the reconstruction and rendering of reflective glass façades. In contrast to 3D Gaussian Splatting (3DGS) [11], which models scenes through volumetric Gaussian ellipsoids, 2DGS leverages flat Gaussian elliptical disks as its scene representation. This disk-based formulation enhances multi-view geometric consistency with respect to both surface normal and depth estimates, enabling reliable extraction of façade surface geometry across varying camera viewpoints. For background regions falling outside the façade masks, standard 2DGS reconstruction and rendering are performed. For regions within the façade masks that correspond to the reflective glass façade, the following specialized procedures are applied.

Each standard 2D Gaussian primitive is defined by the following parameters: a center position $\mathbf{p}_k$, two principal tangential vectors $\mathbf{t}_u$and $\mathbf{t}_v$, two corresponding scaling factors $s_u$and $s_v$, an opacity value $\alpha$, and a spherical harmonics-based color coefficient vector $\mathbf{c}$. In addition to this standard parameterization, we augment each Gaussian primitive with three additional attributes: a reflectivity coefficient $r$, panel semantics $s$ and a 3-dimensional Fresnel base reflectance vector $\mathbf{f}_0$ for computing the Fresnel term. The Fresnel term governs the intensity of specular reflection as a function of the incident viewing angle, effectively reproducing the physically grounded optical behavior of glass, whereby reflectance increases at grazing incidence and decreases when the viewing direction approaches the surface normal.

Within the local tangent plane established in world coordinates, the 2D Gaussian primitive is expressed through Eq. (11) and Eq. (12). For a given point with local coordinates $\mathbf{u} = (u, v)$ on the tangent plane, the corresponding Gaussian evaluation is obtained via Eq. (13).

$$P(u, v) = \mathbf{p}_k + s_u\mathbf{t}_u u + s_v\mathbf{t}_v v = H(u, v, 1,1)^\top \tag{11}$$

$$H = \begin{bmatrix} s_u\mathbf{t}_u & s_v\mathbf{t}_v & 0 & p_k \\ 0 & 0 & 0 & 1 \end{bmatrix} = \begin{bmatrix} RS & \mathbf{p}_k \\ 0 & 1 \end{bmatrix} \tag{12}$$

$$\mathcal{G}(\mathbf{u}) = \exp\left(-\frac{u^2 + v^2}{2}\right) \tag{13}$$

Building upon the above formulations, the rendering process in 2DGS is essentially equivalent to determining the mapping between a camera pixel $\mathbf{x} = (x, y)$ and the local tangent-plane coordinates $\mathbf{u} = (u, v)$ of each 2D Gaussian primitive. This mapping is resolved through a ray–splat intersection operation, where $W \in \mathbb{R}^{4\times4}$represents the world-to-screen transformation matrix:

$$\mathbf{x} = (xz, yz, z, 1)^\top = WH(u, v, 1,1)^\top \tag{14}$$

Once the correspondence from pixel $(x, y)$ to local coordinates $(u, v)$ has been established, the Gaussian contribution at that pixel is evaluated through Eq. (13), and the intersection depth $z$ is derived from Eq. (14). The pixel spherical harmonics-based color $\mathbf{c}(\mathbf{x})$, the reflectivity map $R(\mathbf{x})$, panel semantic map $S(\mathbf{x})$ and the Fresnel base reflectance map $F_0(\mathbf{x})$ are subsequently rendered via volumetric alpha-blending during rasterization:

$$\mathbf{c}(\mathbf{x}) = \sum_{i=1} \mathbf{c}_i \alpha_i \hat{\mathcal{G}}_i(\mathbf{u}(\mathbf{x})) \prod_{j=1}^{i-1} (1 - \alpha_j \hat{\mathcal{G}}_j(\mathbf{u}(\mathbf{x}))) \tag{15}$$

$$R(\mathbf{x}) = \sum_{i=1} r_i \alpha_i \hat{\mathcal{G}}_i(\mathbf{u}(\mathbf{x})) \prod_{j=1}^{i-1} (1 - \alpha_j \hat{\mathcal{G}}_j(\mathbf{u}(\mathbf{x}))) \tag{16}$$

$$S(\mathbf{x}) = \sum_{i=1} s_i \alpha_i \hat{\mathcal{G}}_i(\mathbf{u}(\mathbf{x})) \prod_{j=1}^{i-1} (1 - \alpha_j \hat{\mathcal{G}}_j(\mathbf{u}(\mathbf{x}))) \tag{17}$$

$$F_0(\mathbf{x}) = \sum_{i=1} \mathbf{f}_{0,i} \alpha_i \hat{\mathcal{G}}_i(\mathbf{u}(\mathbf{x})) \prod_{j=1}^{i-1} (1 - \alpha_j \hat{\mathcal{G}}_j(\mathbf{u}(\mathbf{x}))) \tag{18}$$

where $\hat{\mathcal{G}}$ represents a low-pass filtered variant of the Gaussian kernel, which is incorporated to alleviate optimization instabilities arising from highly oblique viewing angles.

During the rendering process, different appearance components are first rendered into separate 2D image maps independently and then composited, following a deferred shading strategy. Specifically, we define a rendered image $I$ as the composition of a diffuse color component calculated by Eq. (15), and a specular reflection component. The diffuse color is represented by the rendered spherical harmonics-based color map, while the specular reflection is estimated through the Reflection MLP detailed in Section 3.3.3. The deferred shading function is formally defined in Eq. (19).

$$I(\mathbf{x}) = (1 - R(\mathbf{x})) \cdot \mathbf{c}(\mathbf{x}) + R(\mathbf{x}) \cdot F(\mathbf{x}) \cdot \mathrm{ReflMLP}(\mathbf{X}, \omega_r) \tag{19}$$

where $\mathbf{X}$ denotes the 3D surface position corresponding to pixel $\mathbf{x}$, and $\omega_r$ represents the reflection direction, which is computed from the viewing direction $\omega_o$ and the local surface normal $\mathbf{n}$ via Eq. (20). Given that reflective glass façades exhibit negligible surface roughness and can thus be treated as ideal specular reflectors, the Fresnel term $F$ is evaluated via Schlick's approximation, as formulated in Eq. (21).

$$\omega_r = \omega_o - 2(\omega_o^\top \mathbf{n})\mathbf{n} \tag{20}$$

$$F = F_0 + (1 - F_0)(1 - \mathbf{n}^\top(-\omega_o))^5 \tag{21}$$

### 3.3.3 Reflection MLP

To achieve accurate modeling of high-frequency near-field specular reflections from observations, we design a Reflection MLP to estimate the specular reflection component. Its detailed architecture is illustrated in Fig. 10. The network architecture consists of eight fully connected layers, each containing 512 hidden units with ReLU activation functions, terminated by an output layer with sigmoid activation. As standard MLPs are inherently biased toward learning low-frequency components [57], while specular reflections are characterized by high-frequency spatial variations, the inputs $\mathbf{X}$ and $\omega_r$ are first projected into a high-dimensional space through Fourier feature encoding [58], as specified in Eq. (22). Before applying the encoding, $\mathbf{X}$ is rescaled to the interval [-1, 1], and $\omega_r$ is constrained to unit length. We adopt $L = 12$ for both $\gamma(\mathbf{X})$ and $\gamma(\omega_r)$ to provide adequate frequency bandwidth for capturing high-frequency specular details.

$$\gamma(p) = (\sin(2^0\pi p), \cos(2^0\pi p), \cdots, \sin(2^{L-1}\pi p), \cos(2^{L-1}\pi p)) \tag{22}$$

The encoded position $\gamma(\mathbf{X})$ is initially processed through the first four layers independently, without being concatenated with the direction input at the outset, allowing the network to capture position-dependent material properties and extract spatially relevant features. This staged fusion strategy facilitates the disentanglement of spatially varying characteristics from view-dependent reflective appearance. At the fifth layer, the learned positional features are combined with the encoded reflection direction $\gamma(\omega_r)$, and $\gamma(\mathbf{X})$ is reintroduced through a skip connection. The skip connection promotes stable gradient propagation and retains fine-grained positional information, alleviating over-smoothing artifacts in deeper layers. The output is ultimately passed through a sigmoid activation to yield the specular reflected RGB color $\mathrm{ReflMLP}(\mathbf{X}, \omega_r)$.

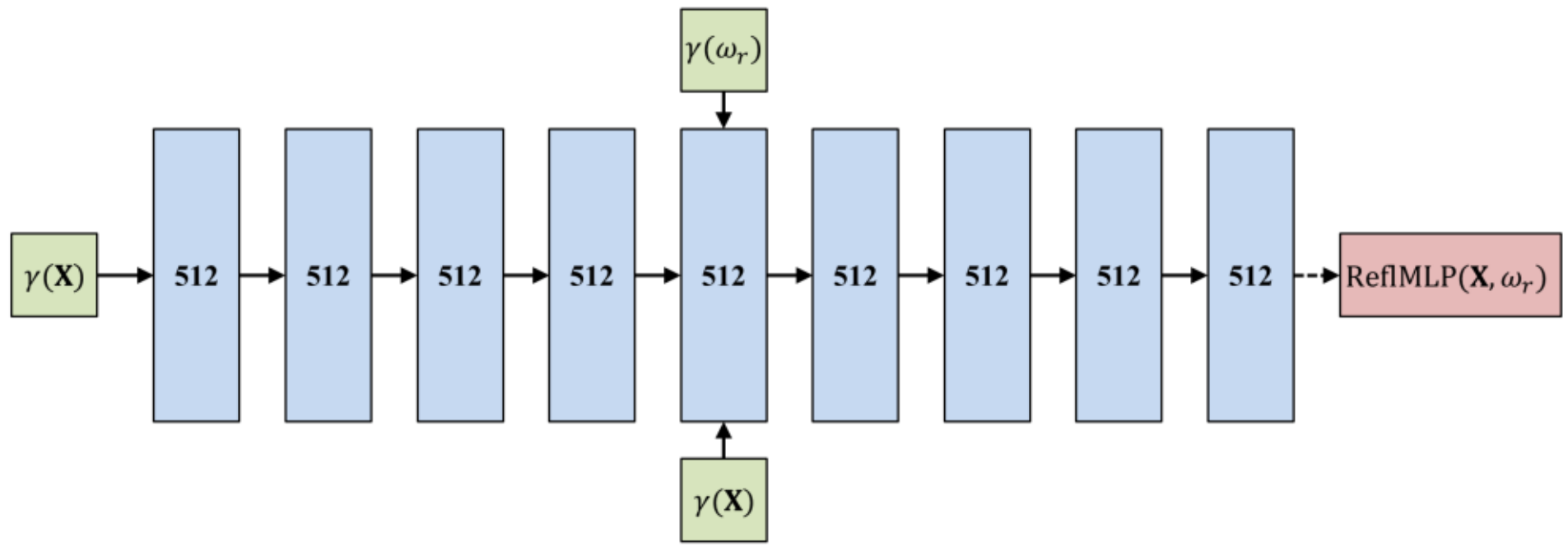

Fig. 10. Architecture of Reflection MLP

#### 3.3.4 Regularization

To fully utilize the precise façade geometry obtained in Section 3.3.1 and to ensure that high-frequency appearance characteristics are adequately captured, we introduce two supplementary regularization terms: a normal-prior loss and a high-frequency loss.

**Normal-prior loss.** As expressed in Eq. (23), we compute an $\mathcal{L}_1$ loss between the rendered surface normals $\mathbf{n}$ and the reference normals $\hat{\mathbf{n}}$ derived from the accurate façade geometry under the corresponding viewpoint. This term enforces the geometric consistency of the reconstructed façade surface.

$$\mathcal{L}_{np} = \frac{1}{|\Omega|}\sum_{\mathbf{x}\in\Omega}\|\mathbf{n}(\mathbf{x}) - \hat{\mathbf{n}}(\mathbf{x})\|_1 \tag{23}$$

where $\Omega$ denotes the set of valid pixels.

**High-frequency loss.** As specified in Eq. (24), we concentrate the supervision signal on pixels whose color residual exceeds a predefined threshold $\tau$. In particular, we identify the subset of pixels exhibiting large $\mathcal{L}_1$ residuals between the rendered color $\mathbf{c}$ and the ground-truth color $\mathbf{c}^{\mathbf{gt}}$, and compute the mean residual over this subset. In practice, we observe that relying solely on the standard RGB reconstruction loss $\mathcal{L}_c$ in Gaussian Splatting tends to spread errors uniformly across the image, producing overly smooth results that fail to faithfully recover sharp, high-frequency reflective details. The proposed high-frequency loss explicitly targets and corrects these high-residual regions.

$$\mathcal{L}_{hf} = \frac{\sum_{\mathbf{x}\in\Omega}\mathbb{I}(\|\mathbf{c}(\mathbf{x}) - \mathbf{c}^{\mathbf{gt}}(\mathbf{x})\|_1 > \tau)\,\|\mathbf{c}(\mathbf{x}) - \mathbf{c}^{\mathbf{gt}}(\mathbf{x})\|_1}{\sum_{\mathbf{x}\in\Omega}\mathbb{I}(\|\mathbf{c}(\mathbf{x}) - \mathbf{c}^{\mathbf{gt}}(\mathbf{x})\|_1 > \tau) + \varepsilon} \tag{24}$$

where $\mathbb{I}(\cdot)$ denotes the indicator function, $\varepsilon$ is a small constant introduced to prevent division by zero, and $\tau$ is initialized at 0.2 and progressively decreased in a stepwise manner over the course of training, reaching a final value of 0.09.

**Overall objective.** The complete training objective is formulated in Eq. (27). Here, $\mathcal{L}_c$ combines an $\mathcal{L}_1$ term with the D-SSIM term, and $\mathcal{L}_n$ denotes the normal-consistency loss; both adhere to the definitions in the original 2DGS formulation, as shown in Eqs. (25) and (26).

$$\mathcal{L}_c = (1-\lambda)\mathcal{L}_1 + \lambda\mathcal{L}_{D-SSIM} \tag{25}$$

where $\lambda = 0.2$.

$$\mathcal{L}_n = \sum_i \omega_i(1 - \mathbf{n}_i^\top\mathbf{N}) \tag{26}$$

where $\mathbf{n}_i$ is the normal of the $i$-th splat, $\omega_i$ is the alpha-blending weight, and $\mathbf{N}$ is the normal estimated from the depth map gradient.

$$\mathcal{L} = \mathcal{L}_c + \lambda_n\mathcal{L}_n + \lambda_{np}\mathcal{L}_{np} + \lambda_{hf}\mathcal{L}_{hf} \tag{27}$$

where $\lambda_n = 0.05$, $\lambda_{np} = 1$, and $\lambda_{hf}$ is initialized at 1 and progressively increased in a stepwise manner during training, reaching a final value of 10.

### 3.4 Object-based Standardized Data Organization for Interactive Facility Management

For the photorealistic and semantic scenes containing reflective glass façades reconstructed via Gaussian Splatting in Section 3.3, the storage format is based on the PLY format, which encodes discrete point data augmented with per-point attributes. While this format is adequate for visualization purposes, facility management, asset documentation, and user interaction demand object-based data organization, similar to the representation employed in BIM models [59,60]. Accordingly, this paper proposes, for the first time, a mapping from the discrete Gaussian primitives of Gaussian Splatting to an object-based representation, thereby rendering the reconstructed scene more readily applicable to real-world interactive applications for facility management.

For a scene reconstructed via Gaussian Splatting, we define three complementary organizational structures: a semantic organization, a non-semantic organization, and an auxiliary rendering organization. The semantic organization structures data into a hierarchical format comprising four layers: *GsScene* Layer, *GsComponent* Layer, *GsElement* Layer, and *GsGaussian* Layer. These layers respectively represent the entire scene, a specific category or building component, the individually segmented elements within a component, and the collection of Gaussian primitives together with their associated attributes that constitute each element. For portions of the scene that have not undergone semantic enhancement, the non-semantic organization is adopted, consisting of three layers: *GsScene* Layer, *GsComponent* Layer, and *GsGaussian* Layer. In addition, the auxiliary rendering organization accommodates supplementary components that assist in rendering. For instance, the network weights of MLPs employed during rendering are stored within this structure. This auxiliary organization is further extensible, allowing the storage of other scene-level information beyond Gaussian primitives, such as environment map representations.

For the RefGlass-GS model generated by our framework, the reflective glass façade portion is mapped to the four-layer semantic organization, where the *GsComponent* Layer represents the entire reflective glass façade, the *GsElement* Layer encodes each individual glass panel, and the *GsGaussian* Layer organizes the augmented 2D Gaussian disks introduced in Section 3.3.2. The background portion is mapped to the three-layer non-semantic organization, where the *GsComponent* Layer represents the entire background and the *GsGaussian* Layer stores standard 2D Gaussian disks. Within the auxiliary rendering organization, the trained weights of the reflective MLP are stored.

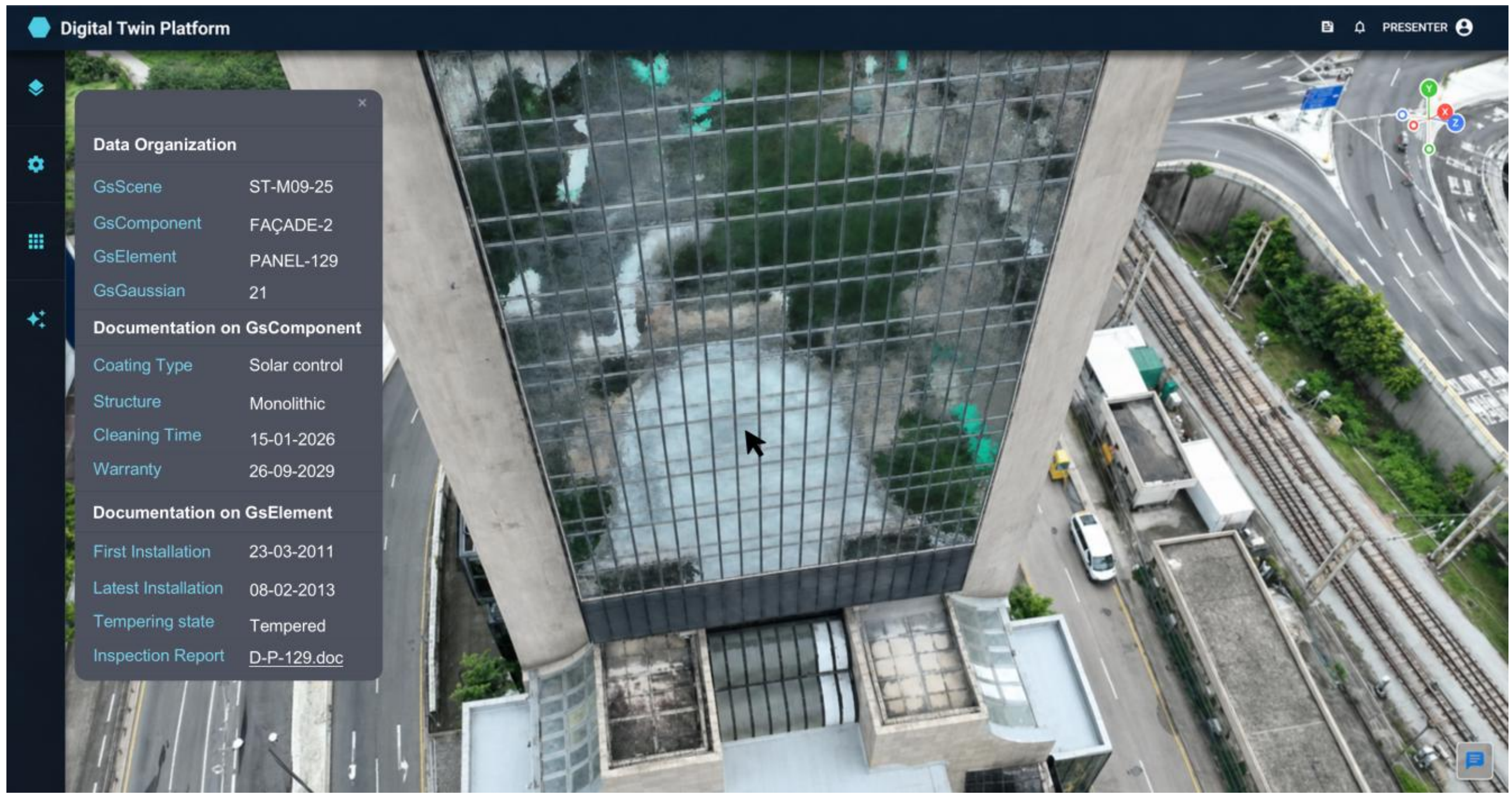

Fig. 11. Interactive facility management of RefGlass-GS model in digital twin platform

The model file organized through the proposed object-based standardized data structure is imported into an interactive digital twin platform [61], as shown in Fig. 11. Scene rendering is performed via Eq. (19). The structured model file facilitates efficient parsing by the platform and enables users to interactively select and inspect individual elements, such as a specific glass panel. Furthermore, additional facility management-related information can be attached at the *GsComponent* Layer or *GsElement* Layer, including material properties, inspection timestamps, defect detection results, and maintenance records. In Fig. 11, we provide hypothetical attributes in the documentation as illustrative examples. This design allows the Gaussian Splatting model to fully leverage its photorealistic scene representation and rendering capabilities while simultaneously fulfilling the functional requirements of practical industry applications.

# 4 Experimental Validation

To validate the effectiveness and superiority of the proposed end-to-end fusion framework for reflective glass façade digitization, we conduct experiments on two real-world reflective glass façade scenes in Hong Kong: an office building located in Sha Tin and the library at The Hong Kong University of Science and Technology, denoted as ST and UST respectively, as shown in Fig. 12. The ST scene features a planar reflective glass façade, whereas the UST scene presents a curved one, thereby allowing the framework to be evaluated under distinct geometric configurations. In the experiments, a DJI M3E UAV equipped with an RTK module is employed for automated image acquisition along the designed viewpoints. The captured images have a resolution of 5280×3956 pixels. The RTK module enables centimeter-level positioning accuracy for the UAV. All experiments presented in this paper are executed on a server equipped with two NVIDIA RTX 4090 GPUs providing a total of 48 GB of memory.

To systematically validate the effectiveness of each proposed method, the following experiments are conducted: Section 4.1 compares the proposed MAP estimation-based glass panel segmentation method exploiting structural regularities against existing deep learning-based glass segmentation approaches, thereby demonstrating its effectiveness in individual glass panel segmentation; Section 4.2 compares the proposed maximal texture coverage-based UAV view planning method with the prevailing commercial nap-of-the-object view planning strategy, thereby demonstrating the effectiveness of the proposed data acquisition scheme in recovering view-dependent appearance; Section 4.3 presents baseline comparisons, ablation studies, and validation of the Reflection MLP design for the proposed RefGlass-GS modeling method.

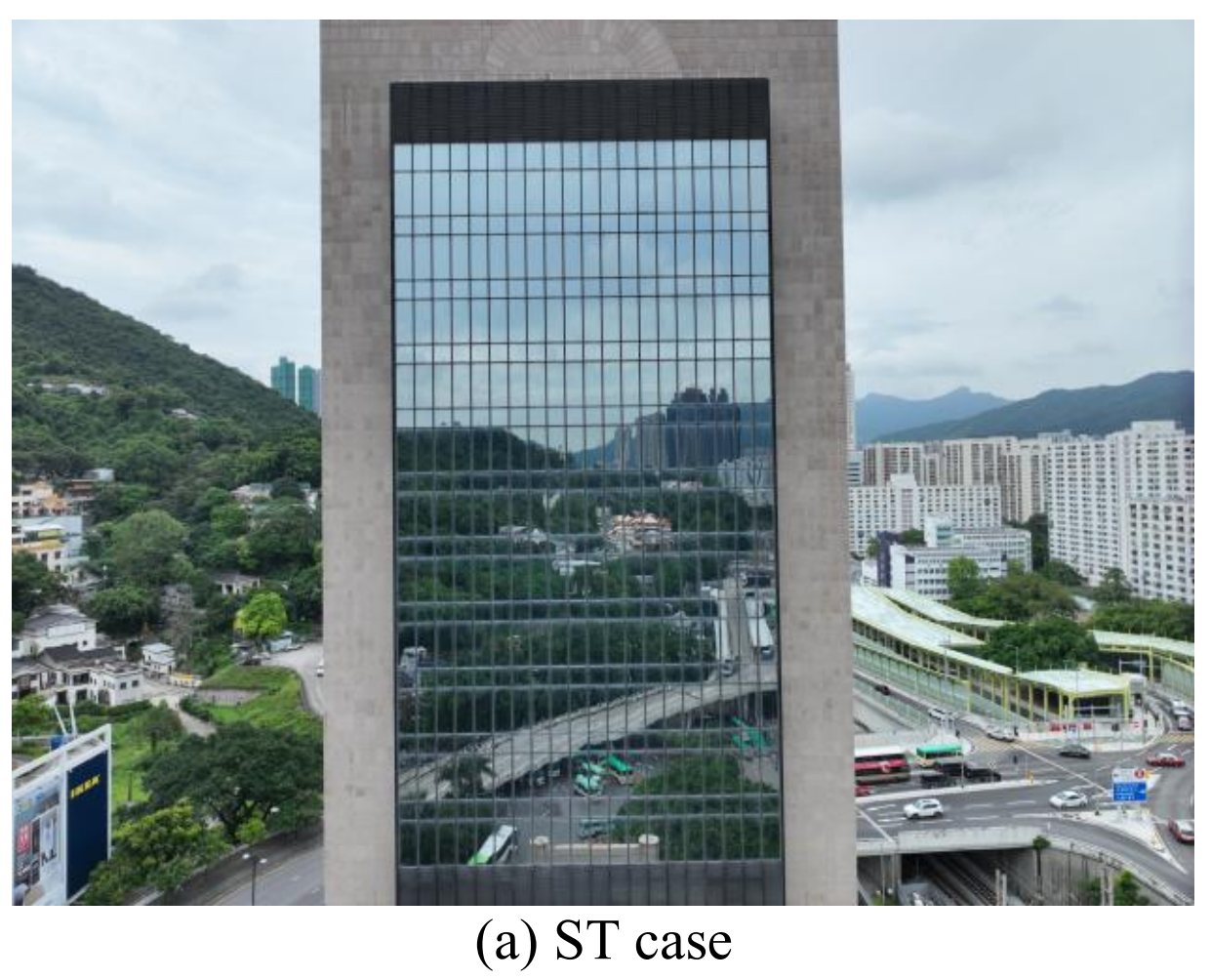

(a) ST case

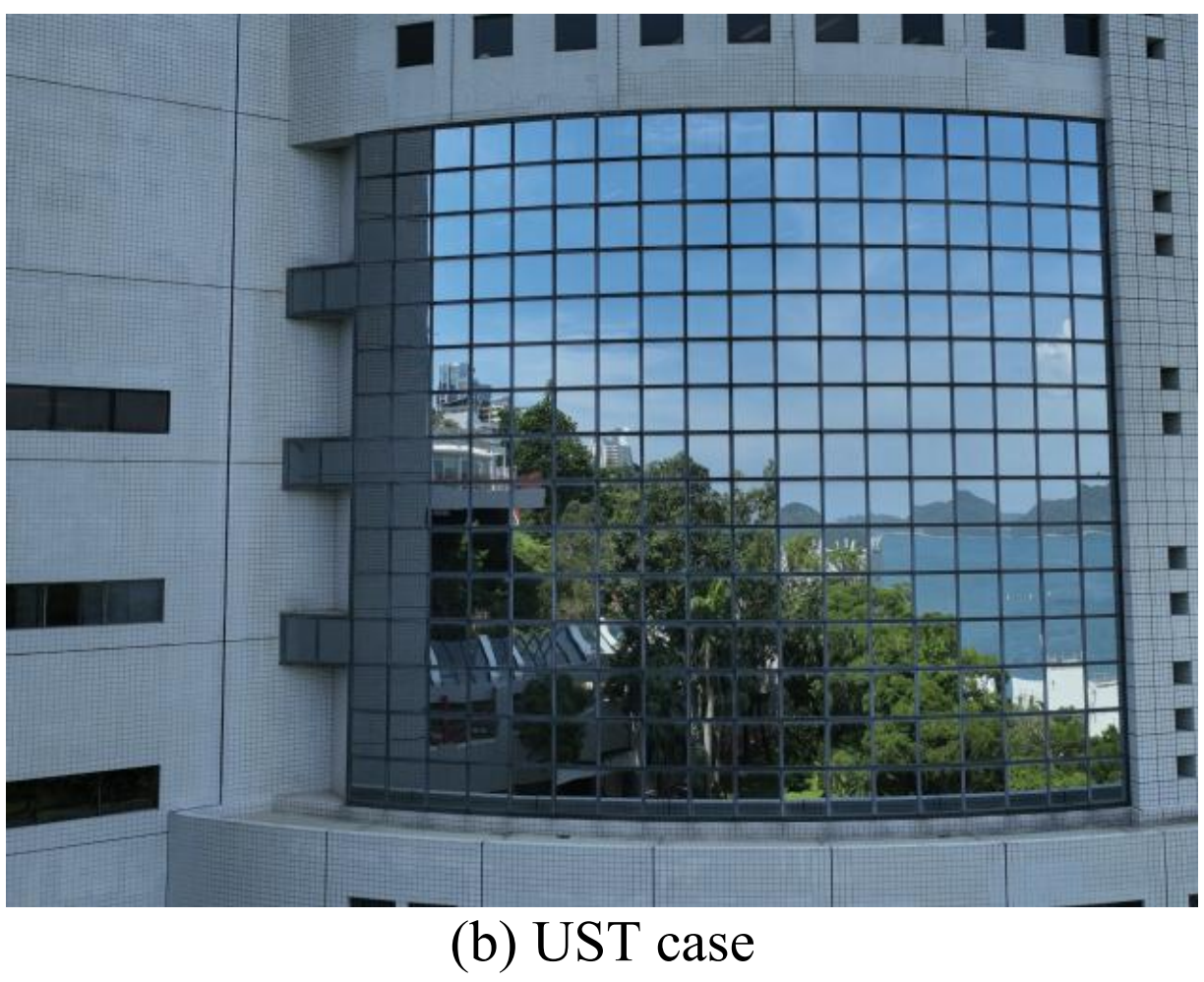

(b) UST case

Fig. 12. Illustration of two validation scenes

## 4.1 Evaluation on Glass Panel Segmentation

We first validate the proposed individual glass panel segmentation method applied to entire glass façades. Three state-of-the-art deep learning-based glass segmentation methods are selected as baselines: EBLNet [6], GhostingNet [7], and GlassSemNet [8]. The images used for validation are those captured during the second flight described in Section 3.3.1. The segmentation masks produced by our method are obtained by back-projecting the 3D glass panels derived in Section 3.1.2, which originate from the MAP estimation-based individual glass panel segmentation of a single fronto-parallel view image in Section 3.1, onto each image. These back-projection images allow the evaluation to extend beyond the single fronto-parallel view used in Section 3.1, encompassing a broader set of multi-view images for glass panel segmentation assessment.

The quantitative evaluation metrics are organized into three categories. The first category computes the overall mean Intersection-over-Union (mIoU) between the predicted glass mask and the ground-truth reflective glass façade region, measuring whether the entire reflective glass façade area is correctly distinguished from the remaining materials. The second category evaluates the coverage of the predicted masks over individual glass panels; specifically, for each panel, the coverage ratio is first computed, and then the 20th, 10th, and 5th percentiles across all panel coverage values are reported, thereby indicating whether any individual glass panels are missed during segmentation and quantifying the severity of such omissions. The third category calculates the mIoU between the predicted individual glass panel masks and the corresponding ground-truth individual panel masks, directly assessing the effectiveness and accuracy of the method in individual glass panel segmentation. Both ground-truth overall glass façade masks and individual panel masks are obtained through careful manual annotation on the captured RGB images.

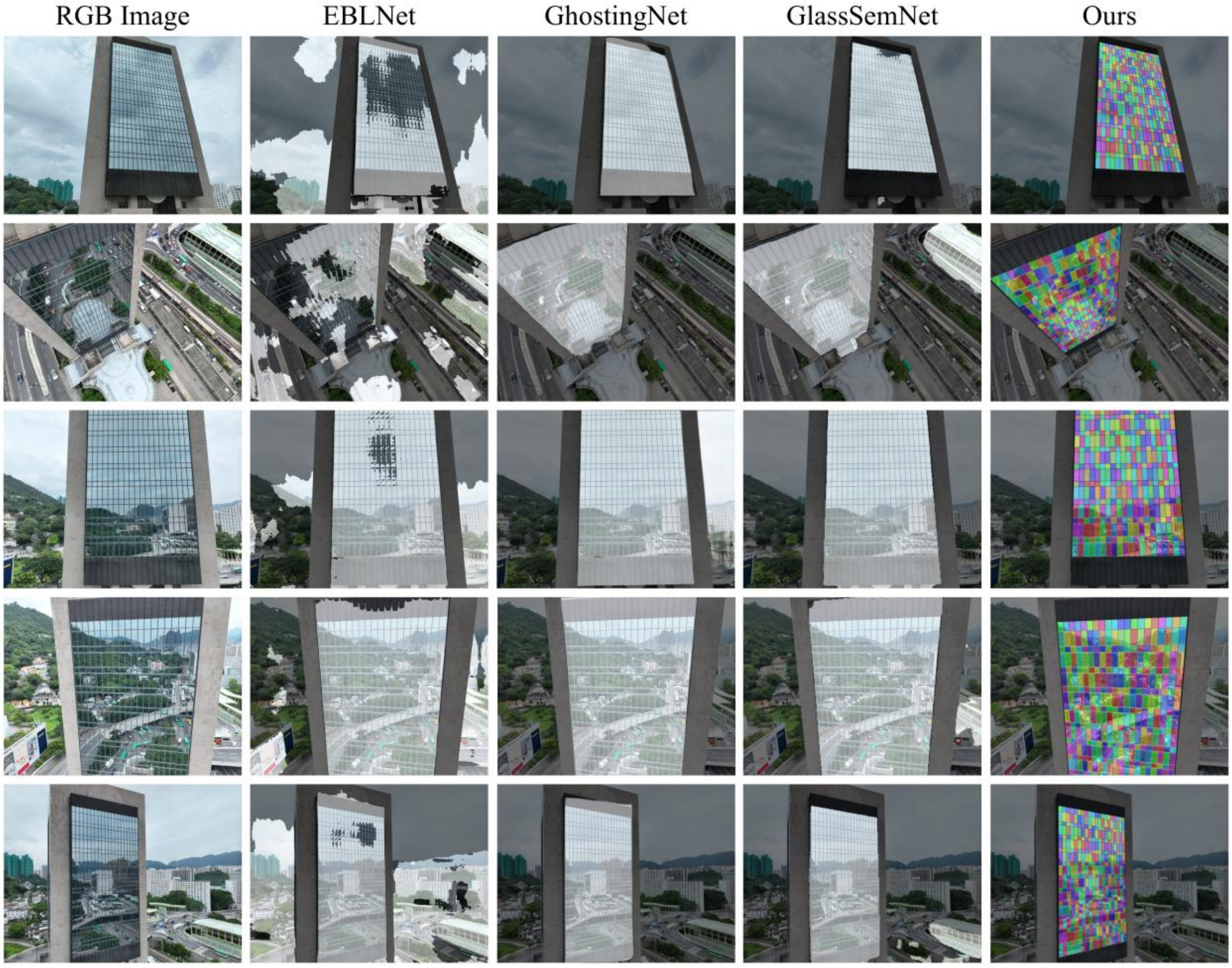

(a) Glass segmentation on ST case

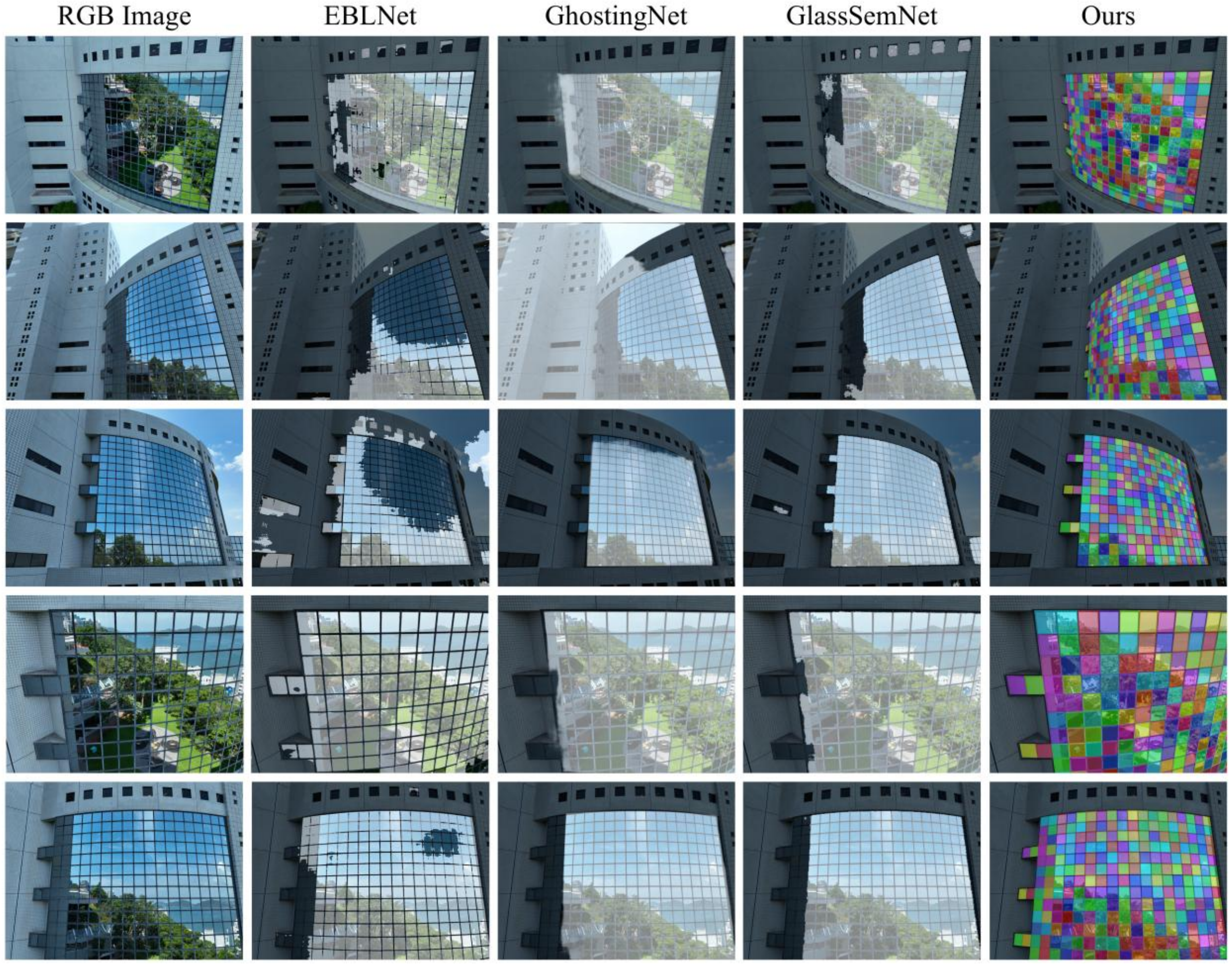

(b) Glass segmentation on UST case

Fig. 13. Qualitative comparison of glass panel segmentation. For EBLNet, GhostingNet, and GlassSemNet, white masks indicate glass material segmentation, whereas for our method, different colored masks represent distinct individual panels.

The quantitative and qualitative comparison results are presented in Table 1 and Fig. 13, respectively. As evidenced by the results, under the interference of reflective textures and background clutter, our proposed glass panel segmentation method achieves an average improvement of 0.1927 in mIoU over the best-performing SOTA methods in overall glass material distinction. Moreover, only our method is capable of performing individual glass panel segmentation, and the coverage percentiles confirm that our approach does not miss any individual glass panels during segmentation.

Upon further analysis, in the ST case featuring a planar reflective glass façade, EBLNet and GlassSemNet are susceptible to background interference, erroneously segmenting portions of the background as glass, with EBLNet being more severely affected. Furthermore, due to the interference of reflective textures, EBLNet exhibits segmentation omissions within regions that belong to glass, as reflected in the coverage 5th percentile metric for the ST scene. Although GhostingNet is able to segment the reflective glass façade relatively completely in the ST case, it produces erroneous segmentation at glass boundary regions and other peripheral areas. In the UST case featuring a curved reflective glass façade, EBLNet, GhostingNet, and GlassSemNet all exhibit varying degrees of missed panel segmentation. Compared with the planar glass case, their performance deteriorates in both overall mIoU and coverage percentiles, demonstrating that existing SOTA methods are less effective in segmenting non-planar glass surfaces than planar ones. In contrast, our method overcomes the limitations posed by curved glass geometry as well as the interference from background and reflective textures by exploiting structural regularities, thereby achieving complete and omission-free panel segmentation.

Table 1. Quantitative comparison results of glass panel segmentation

| Metrics | EBLNet [6] | GhostingNet [7] | GlassSemNet [8] | **Ours** |
|---|---|---|---|---|
| ST | | | | |
| Overall mIoU | 0.4701 | 0.8597 | 0.8020 | 0.9895 |
| Coverage 20% percentile | 1.0000 | 1.0000 | 1.0000 | 1.0000 |
| Coverage 10% percentile | 0.7829 | 1.0000 | 1.0000 | 1.0000 |
| Coverage 5% percentile | 0.2881 | 1.0000 | 1.0000 | 1.0000 |
| Individual mIoU | N/A | N/A | N/A | 0.9787 |
| UST | | | | |
| Overall mIoU | 0.4339 | 0.7316 | 0.6218 | 0.9871 |
| Coverage 20% percentile | 0.1128 | 0.7277 | 0.9058 | 1.0000 |
| Coverage 10% percentile | 0.0127 | 0.3126 | 0.1824 | 1.0000 |
| Coverage 5% percentile | 0.0031 | 0.1641 | 0.0217 | 1.0000 |
| Individual mIoU | N/A | N/A | N/A | 0.9709 |

### 4.2 Evaluation on UAV View Planning

We further conduct experimental validation on the UAV view planning strategy for maximal texture coverage proposed in Section 3.2, examining whether it outperforms existing commonly used UAV view planning methods designed for high-precision geometric reconstruction in terms of view-dependent texture recovery. Prior to planning, the number of UAV viewpoints required for each case must first be determined; this parametric analysis is presented in Section 4.2.1. Subsequently, with the number of viewpoints established, Section 4.2.2 compares the RefGlass-GS reconstruction results obtained using our view planning method against those obtained using nap-of-the-object view planning, which has demonstrated strong performance in geometric reconstruction, thereby validating the effectiveness of our approach.

#### 4.2.1 Parametric Analysis

Since the number of viewpoints is a prior parameter required as input for the optimization in Section 3.2, we conduct a parametric analysis on the viewpoint number to ensure appropriate and efficient image capturing. For both the ST and UST cases, we vary the number of viewpoints from 100 to 800 with increments of 20, solving the optimization formulation in Eq. (9) with each viewpoint number as input. The most suitable number of viewpoints for each case is then determined by evaluating the marginal gain in the final fitness value as the number of viewpoints increases.

Fig. 14 illustrates the results of the parametric analysis on viewpoint number for the ST and UST cases. To account for the randomness inherent in the optimization process, we run the optimization three times for each viewpoint number in each case and take the average, which is displayed as the bold black line. Additionally, we apply a degree of smoothing to the viewpoint number versus fitness value curve to better identify the marginal effect of increasing the viewpoint number. As shown in Fig. 14, the optimized fitness value increases with the number of viewpoints for both the ST and UST cases. For the ST case, the improvement in fitness value is relatively pronounced before 560 viewpoints and gradually levels off thereafter. For the UST case, 600 viewpoints marks the point at which the optimized fitness value begins to plateau. Based on this parametric analysis, we select 560 viewpoints for the ST case and 600 for the UST case as the optimization prior input for determining the viewpoints used in second flight image capturing.

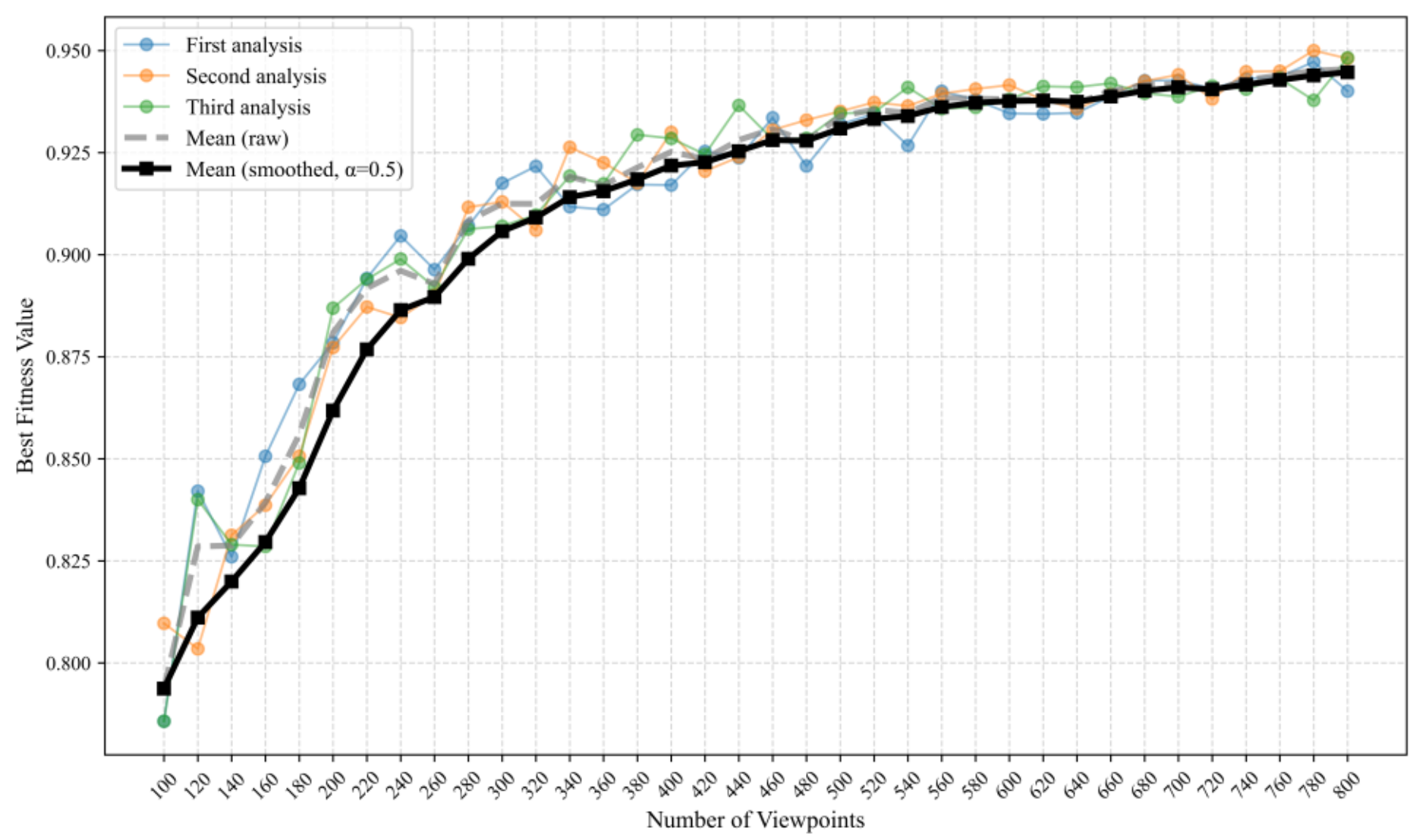

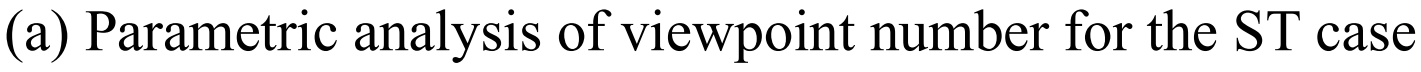
(a) Parametric analysis of viewpoint number for the ST case

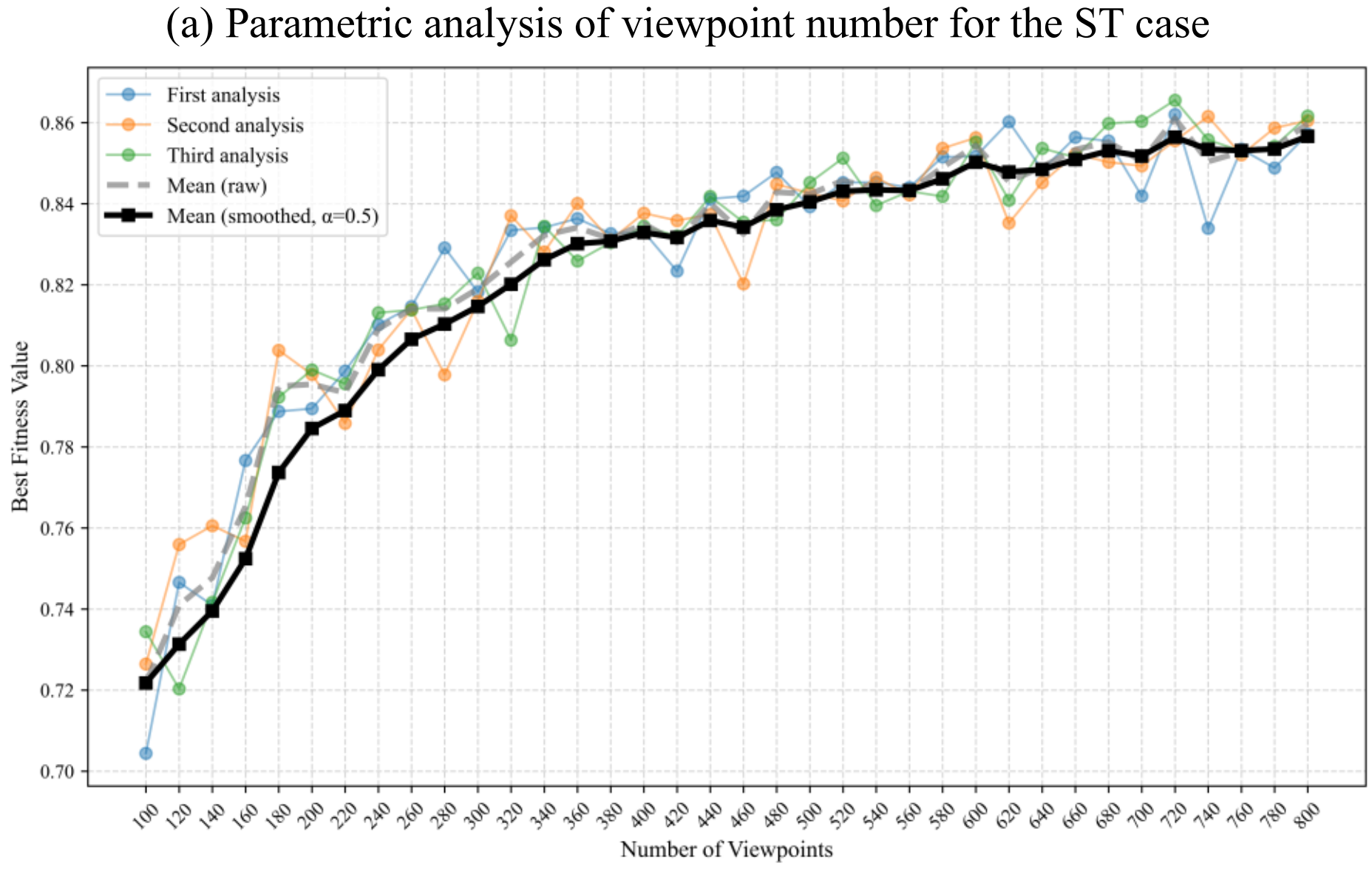

(b) Parametric analysis of viewpoint number for the UST case

Fig. 14. Parametric analysis of UAV viewpoint number

### 4.2.2 Comparison with View Planning by Nap-of-the-object

In high-precision geometric reconstruction, nap-of-the-object (NOTO) view planning is an effective and commercially adopted method. Unlike two-dimensional orthophoto and three-dimensional oblique photogrammetry, NOTO plans UAV flight paths that closely follow the object's surface contour [62]. In our implementation, NOTO view planning is realized through the Slope Mission Plan of the DJI M3E for automated flight planning.

We compare the two view planning methods with the same 600 viewpoints in the UST case, as illustrated in Fig. 15. Automated UAV image acquisition is conducted for both our maximal texture coverage view planning and NOTO view planning within the same time period of another test. The glass façade regions from both datasets are then extracted via 3D glass panel back-projection, and both sets of images are subsequently modeled using the RefGlass-GS method proposed in Section 3.3. Finally, the two resulting GS models are rendered at novel views that are excluded from the training set to evaluate the capability of each view planning strategy in recovering view-dependent textures. We adopt the commonly used PSNR, SSIM [54], and LPIPS [63] as comparison metrics.

The quantitative and qualitative comparison results are presented in Table 2 and Fig. 16, respectively. As shown by the results, our view planning substantially outperforms NOTO in view-dependent texture recovery, achieving an improvement of 13.15 dB in PSNR. As can be observed from Fig. 16, the data captured by NOTO suffers from severe blurring during texture rendering due to the lack of consideration for dynamic textures. In contrast, our maximal texture coverage strategy is capable of recovering highly fine-grained and detailed view-dependent textures.

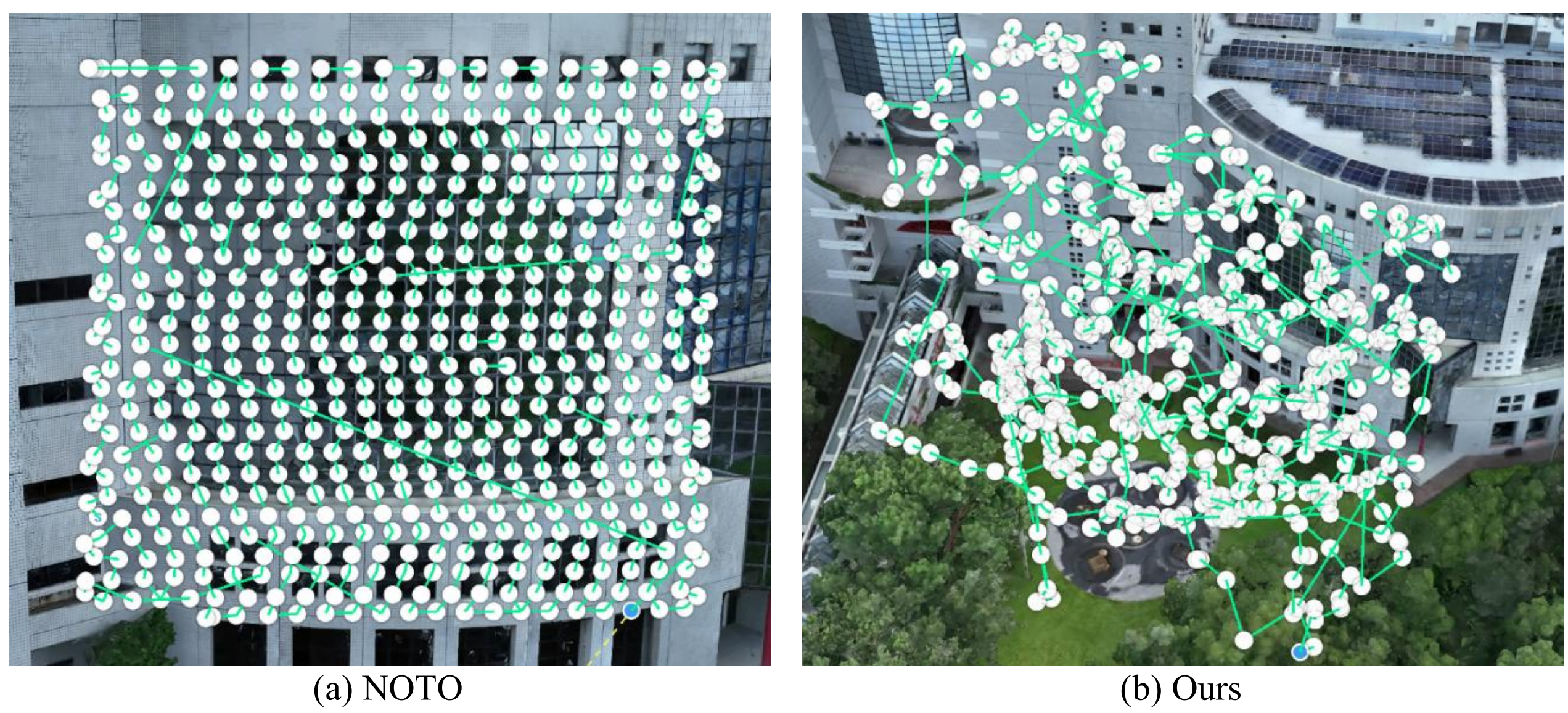

(a) NOTO (b) Ours

Fig. 15. Illustration of two UAV view planning strategies

Table 2. Quantitative comparison results of UAV view planning strategies

| UAV view planning mode | PSNR↑ | SSIM↑ | LPIPS↓ |
| --- | --- | --- | --- |
| Nap-of-the-object (NOTO) | 18.44 | 0.7031 | 0.3006 |
| **Maximal texture coverage (Ours)** | 31.59 | 0.9247 | 0.0701 |

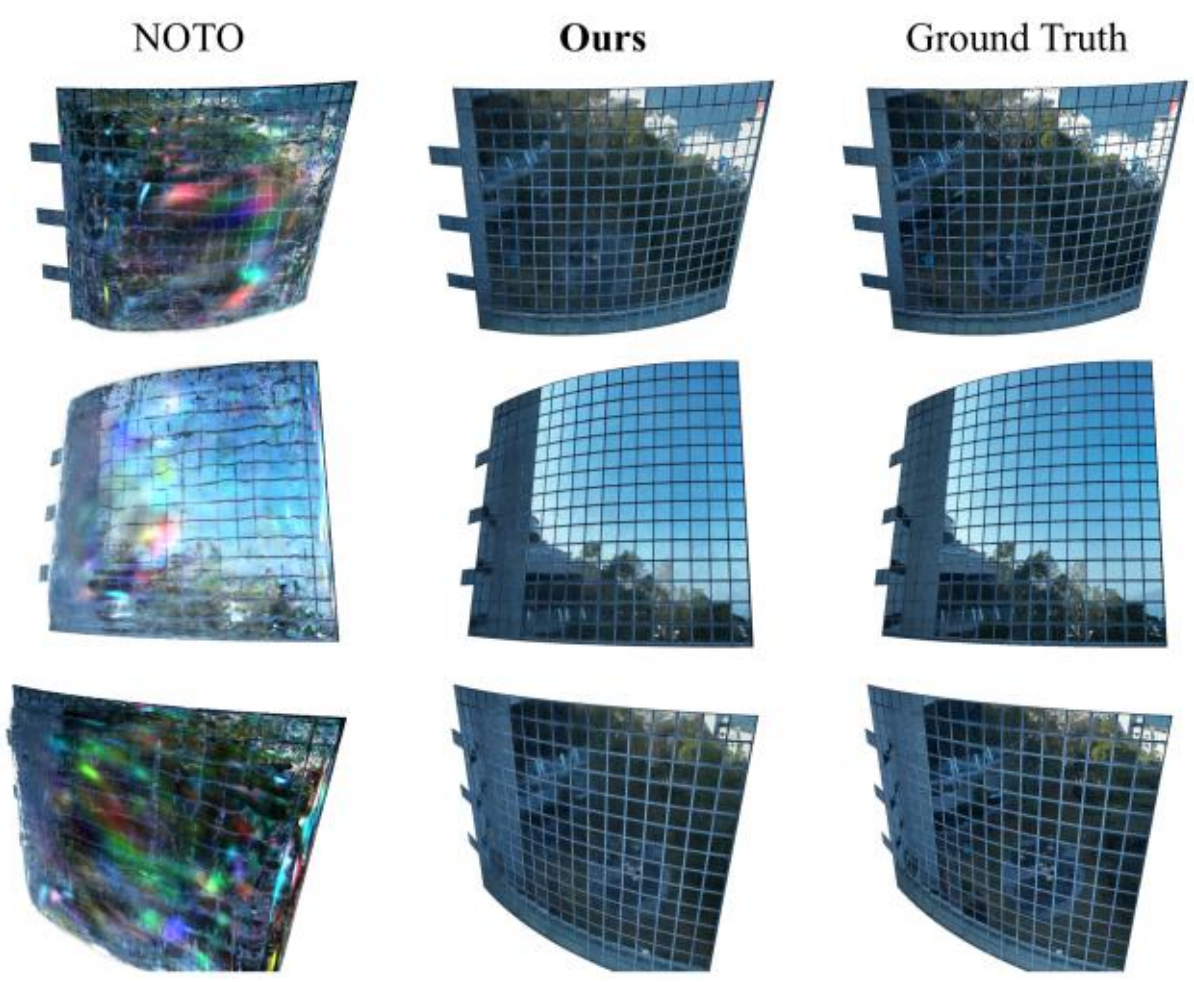

Fig. 16. Qualitative comparison results of UAV view planning strategies

## 4.3 Evaluation of RefGlass-GS Modeling

For the RefGlass-GS method of reflective glass façade modeling proposed in Section 3.3, we conduct experimental validation from three aspects. First, in Section 4.3.1, we perform a baseline comparison against SOTA Gaussian Splatting methods for reflective scene reconstruction and rendering to demonstrate its superiority. Second, in Section 4.3.2, we explore and compare different architecture designs of the Reflection MLP, which is the key module responsible for reconstructing and rendering high-frequency near-field appearance in RefGlass-GS, thereby validating the rationality of our design. Additionally, in Section 4.3.3, we conduct an ablation study to verify the necessity of each component.

### 4.3.1 Baseline Comparison

For the baselines used in comparison, we select vanilla 3DGS [11] and 2DGS [20], as well as SOTA Gaussian Splatting methods specifically designed for reflective scenes and reflection recovery, including GaussianShader [13], Ref-Gaussian [14], and Ref-GS [29]. Comparison experiments are conducted on both the ST and UST cases. For the baseline methods, we adopt their publicly available configurations and hyperparameter settings. For our RefGlass-GS, the number of training iterations is set to $2\times10^5$. Similar to Section 4.2.2, we select the reflective glass façade regions for comparison.

Table 3 and Fig. 17 present the quantitative and qualitative results of the baseline comparison, respectively. As shown by the results, our RefGlass-GS achieves superior scene modeling and inverse rendering in real-world reflective glass façade scenarios that exhibit complex high-frequency near-field reflections. In the ST and UST cases, our RefGlass-GS outperforms the existing best method in PSNR by 5.15 dB and 5.01 dB, respectively.

The superiority of our RefGlass-GS in modeling reflective appearance can be further examined in Fig. 17, where the red dashed boxes indicate the fine-grained details successfully recovered by our method. Ref-Gaussian is capable of partially recovering view-dependent appearance; however, it employs a 2D environment map to represent global illumination, which introduces parallax errors when modeling real-world 3D near-field reflections and consequently produces visibly blurred high-frequency details. GaussianShader similarly relies on a 2D environment map but adopts 3D Gaussians as primitives, with normals defined along the minor-axis direction of each 3D Gaussian. Such a representation leads to inaccurate façade-surface geometry, which subsequently results in erroneous reflection estimation. Consequently, it exhibits locally chaotic appearance in the central region of the façade, or shows unstable appearance across extensive façade areas, which is attributable to poor geometric reconstruction. For Ref-GS, although it builds upon deferred rendering of 2D Gaussian Splatting and applies directional encoding to the deferred-rendered surface to model view-dependent effects, its normal estimation fails to accurately recover the geometry of glass façade surfaces. This inaccuracy undermines the effectiveness of its spherical mip-grid based directional encoding in reconstructing the specular reflection effects, thereby also generating significant visual disorder across the façade. Furthermore, vanilla 3DGS and 2DGS rely on spherical harmonics to represent color, which is inherently limited to capturing low-frequency view-dependent variations and proves inadequate for reproducing high-frequency specular reflections.

Table 3. Quantitative comparison results of reflective glass façade modeling

| Model | ST | | | UST | | |
|---|---|---|---|---|---|---|
| | PSNR↑ | SSIM↑ | LPIPS↓ | PSNR↑ | SSIM↑ | LPIPS↓ |
| 3DGS [11] | 21.98 | 0.8566 | 0.1135 | 22.70 | 0.8521 | 0.1278 |
| 2DGS [20] | 21.10 | 0.8474 | 0.1250 | 22.40 | 0.8511 | 0.1346 |
| GaussianShader [13] | 21.40 | 0.8350 | 0.1638 | 22.62 | 0.8402 | 0.1599 |
| Ref-Gaussian [14] | 25.02 | 0.9139 | 0.0797 | 26.65 | 0.9057 | 0.0930 |
| Ref-GS [29] | 22.49 | 0.8462 | 0.1406 | 23.36 | 0.8588 | 0.1399 |
| **RefGlass-GS (Ours)** | 30.17 | 0.9339 | 0.0594 | 31.66 | 0.9237 | 0.0695 |

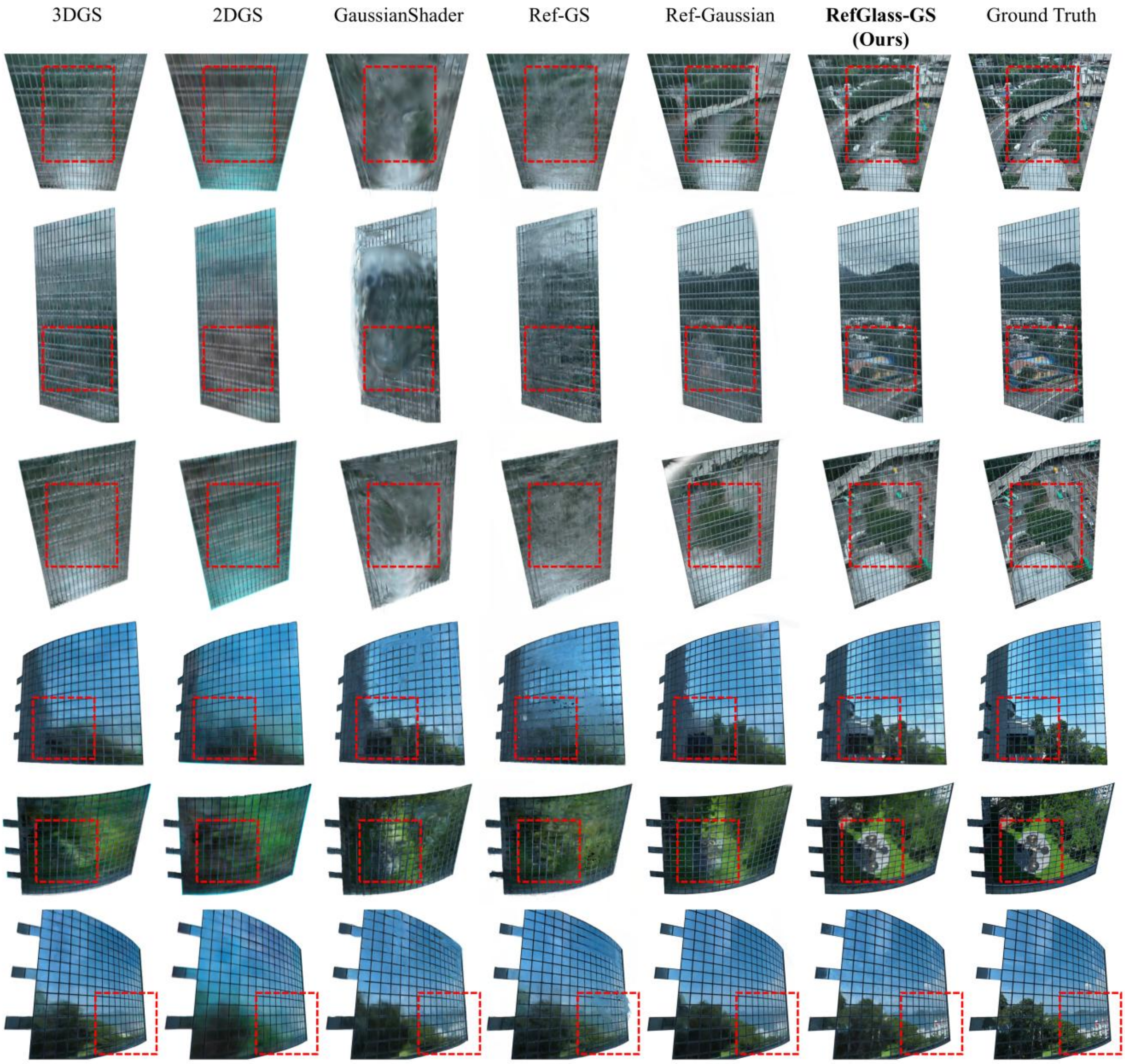

Fig. 17. Qualitative comparison results of reflective glass façades modeling

In summary, the above comparisons confirm that our RefGlass-GS effectively models reflective glass façade using augmented 2D Gaussian primitives and captures near-field specular reflections through a dedicated Reflection MLP, collectively offering a robust solution for high-fidelity digitization of reflective glass façades.

### 4.3.2 Comparison of Reflection MLP Architecture Designs

To validate the rationality of the Reflection MLP architecture designed in Section 3.3.3, we conduct a dedicated ablation study on the architecture of the Reflection MLP. Specifically, five factors influence the architecture of the Reflection MLP: (1) the hidden layer dimension of the MLP, denoted as **Dim**; (2) the network depth of the MLP, denoted as **Layer**; (3) the Fourier feature encoding order applied to the position and reflection direction, denoted as ***L***; (4) whether to incorporate a skip connection that re-injects the encoded position features into the network, denoted as **w/ skip** or **w/o skip**; and (5) whether to adopt the staged fusion strategy that feeds the encoded reflection direction at an intermediate layer of the network rather than concatenating it with the encoded position at the initial input, denoted as **w/ staged** or **w/o staged**. We vary the configurations of these five factors to assess their respective influences on modeling quality.

A variety of Reflection MLP architectures are assessed on the UST case, with results summarized in Table 4. Case 4, corresponding to our final architecture configuration, attains the highest overall performance, confirming the effectiveness of our Reflection MLP design. A comparison among Cases 1, 2, and 4 reveals that enlarging the hidden-layer dimension and increasing the network depth are both beneficial for reconstructing high-frequency, view-dependent textures. The underlying reason is that the Reflection MLP functions as a high-capacity nonlinear approximator mapping spatial position and reflected direction to reflected color; broader layers enhance feature representational power, while additional depth strengthens the capacity to capture complex nonlinear dependencies between location and viewing direction. The comparison between Cases 4 and 6 demonstrates that incorporating a skip connection to re-inject the encoded position features alleviates gradient forgetting throughout the training process, yielding a notable improvement in reflection recovery. The comparison between Cases 4 and 5 indicates that the staged fusion strategy decouples the learning of position-dependent and view-dependent features, thereby improving the accuracy of near-field reflection estimation. Furthermore, Cases 3, 4, 7, and 8 collectively suggest that a moderate increase in the Fourier feature encoding order facilitates the representation of high-frequency components, yet an excessively high encoding order leads to performance degradation, as overly high-frequency basis functions intensify the network's sensitivity to minor pose inaccuracies and pixel-level noise.

Table 4. Quantitative comparison results of Reflection MLP design

| Case | Architecture | PSNR↑ | SSIM↑ | LPIPS↓ |
|---|---|---|---|---|
| 1 | Dim = 256, Layer = 8, $L$ = 12, w/ skip, w/ staged | 31.12 | 0.9198 | 0.0714 |
| 2 | Dim = 512, Layer = 6, $L$ = 12, w/ skip, w/ staged | 31.61 | 0.9232 | 0.0699 |
| 3 | Dim = 512, Layer = 8, $L$ = 10, w/ skip, w/ staged | 31.62 | 0.9231 | 0.0698 |
| 4 | **Dim = 512, Layer = 8, $L$ = 12, w/ skip, w/ staged (Ours)** | 31.66 | 0.9237 | 0.0695 |
| 5 | Dim = 512, Layer = 8, $L$ = 12, w/ skip, w/o staged | 31.51 | 0.9229 | 0.0699 |
| 6 | Dim = 512, Layer = 8, $L$ = 12, w/o skip, w/ staged | 31.24 | 0.9199 | 0.0709 |
| 7 | Dim = 512, Layer = 8, $L$ = 14, w/ skip, w/ staged | 31.39 | 0.9211 | 0.0704 |
| 8 | Dim = 512, Layer = 8, $L$ = 16, w/ skip, w/ staged | 30.84 | 0.9157 | 0.0736 |

#### 4.3.3 Ablation Study

The ablation study targets three component innovations of RefGlass-GS, namely the Reflection MLP, the high-frequency loss, and the normal-prior loss. Experiments are conducted on the UST case. Table 5 and Fig. 18 present the quantitative and qualitative results of the ablation study, respectively. The results show that our complete RefGlass-GS achieves the best performance, validating the effectiveness of all three components.

For the w/o Reflection MLP case in Table 5, we replace the Reflection MLP described in Section 3.3.3 with a standard environment map. The results indicate that the infinite-distance lighting assumption inherent in the environment map introduces noticeable parallax errors in real-world near-field reflection scenarios, whereas our Reflection MLP representation effectively overcomes this limitation.

For the w/o high-frequency loss case in Table 5, we remove the high-frequency regularization term $\mathcal{L}_{hf}$ proposed in Section 3.3.4 and retain only the standard color loss $\mathcal{L}_c$ that averages over the entire image for color supervision. The results show that, without explicit supervision on high-frequency color details, the training signal merely drives global average color improvement across the full image, leading to a degradation in the sharpness of fine-grained high-frequency details.

For the w/o normal-prior loss case in Table 5, we remove the normal-prior regularization term $\mathcal{L}_{np}$ proposed in Section 3.3.4, discarding the prior geometric information and relying solely on color signals for geometry reconstruction. The results reveal that, due to inaccuracies in geometry estimation and the high sensitivity of reflection computation to surface normal orientation, errors are introduced in reflection recovery.

Table 5. Quantitative results of ablation studies

| | PSNR↑ | SSIM↑ | LPIPS↓ |
|---|---|---|---|
| w/o Reflection MLP | 28.25 | 0.9023 | 0.0855 |
| w/o high-frequency loss | 31.13 | 0.9216 | 0.0705 |
| w/o normal-prior loss | 31.28 | 0.9195 | 0.0718 |
| **Ours** | 31.66 | 0.9237 | 0.0695 |

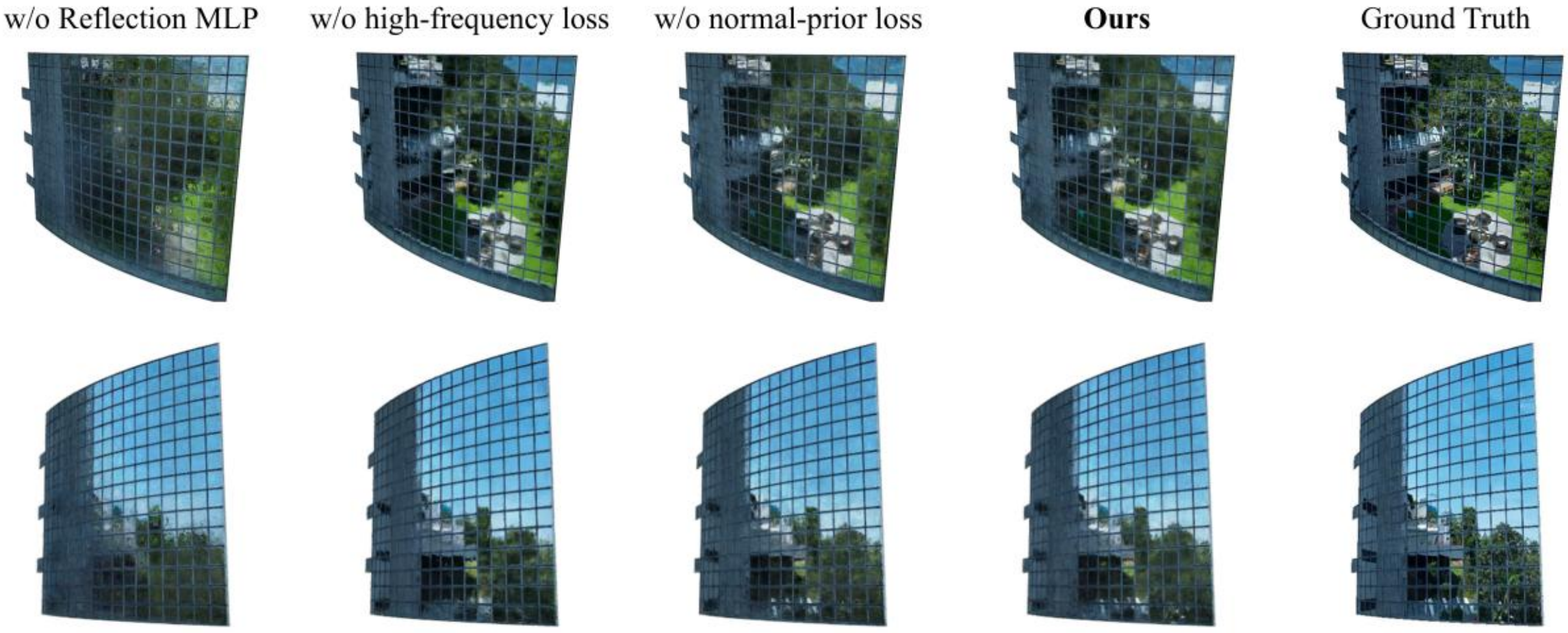

Fig. 18. Qualitative results of ablation studies

## 5 Limitations and Future Work

Although the RefGlass-GS fusion framework proposed in this paper achieves a unification of scientific contribution and industrial application for reflective glass façade digitization, and the extensive experiments in Section 4 have validated the superiority of our algorithmic and methodological innovations, the framework still presents certain limitations in practical applications. First, the glass panel segmentation method in this paper primarily targets rectangular glass panels. While these represent the most commonly encountered panel geometry in urban environments, the method's capability to segment panels of other shapes, such as triangular or irregular forms, remains limited. Second, this paper mainly addresses reflective-dominant glass façades. For transmissive glass façades, where coupled double-layer depths exist at each observation viewpoint [64], the proposed method does not account for the reconstruction of multi-layer geometry.

Future work will focus on extending the glass panel segmentation method to irregular panel geometries, thereby enhancing the generalizability of the framework. In addition, we will extend this semantic, photorealistic, and interactive digitization fusion framework to transmissive glass façades, enabling it to better serve the development of digital cities and smart cities. Furthermore, we intend to integrate physics-based simulation into the digitalized building Gaussian Splatting models [65], expanding their functionality beyond visualization and documentation to realize the vision of "what we can see is what we can simulate."

## 6 Conclusions

This paper proposes an end-to-end UAV-enabled fusion framework based on Gaussian Splatting for reflective glass façade digitization. Driven by four practical requirements, namely (1) individual-level panel semantic segmentation, (2) UAV view planning for efficient and effective data collection, (3) effective modeling of the high-frequency near-field reflections inherent on real-world façade surfaces, and (4) standardized data organization for practical facility management applications and digital twin platform interaction, we develop four corresponding technical innovations that collectively achieve a unified digitization solution encompassing photorealistic, semantic, and interactive dimensions. Regarding the four technical innovations, we first propose a robust individual glass panel segmentation method based on structural regularities maximum a posteriori estimation, enabling reliable panel-level segmentation under the interference of reflective textures and complex backgrounds. Second, we propose a maximal texture coverage UAV planning strategy tailored for objects with view-dependent appearance, which constructs a unit hemisphere physical model and formulates the planning task as an orienteering problem mathematical optimization model to achieve efficient and comprehensive data acquisition. Third, we propose the RefGlass-GS modeling method, which builds upon augmented 2D Gaussian Splatting primitives and achieves high-fidelity reconstruction and rendering of challenging real-world reflective glass façades through a novel deferred shading formulation, a Reflection MLP, and newly developed normal-prior loss and high-frequency loss for regularization. Finally, we construct a general four-layer semantic organization that maps the Gaussian Splatting model from discrete point data to an object-based representation, supporting model loading and user interaction within practical digital twin platforms.

Extensive validation experiments substantiate our contributions and demonstrate their superiority. In terms of glass panel segmentation, we compare our method with three SOTA deep learning based glass segmentation approaches. Our proposed method achieves an average improvement of 0.1927 in mIoU over the best-performing existing method in overall glass façade segmentation, and notably, only our method is capable of performing individual panel-level segmentation without omissions in panel coverage. In terms of UAV view planning, we compare our approach with nap-of-the-object view planning that prioritizes high-precision geometric reconstruction. In the task of novel view synthesis for view-dependent textures, our maximal texture coverage view planning achieves an improvement of 13.15 dB in PSNR. In terms of reflection modeling, we compare RefGlass-GS with five existing Gaussian Splatting modeling methods. In the ST and UST cases, our RefGlass-GS outperforms the best existing method in PSNR by 5.15 dB and 5.01 dB, respectively. In addition, we conduct a comparative exploration of different architecture designs for the Reflection MLP, as well as the ablation study on the three key components of RefGlass-GS, collectively validating the rationality of our model design.

The fusion architecture proposed in this paper encompasses both algorithmic innovation and practical application value, and is expected to further advance Gaussian Splatting toward real-world deployment scenarios in digital twin based smart city applications such as facility management.

**Acknowledgments**

The authors would like to acknowledge the support by RGC Theme-based Research Schemes (2023/24 T22-606/23-R), RGC Theme-based Research Schemes (2024/25 T22-607/24-N), Strategic Public Policy Research Funding Scheme (S2024.A7.022), and RGC Early Career Scheme (Project No. 9048329). The authors declare that they have no known competing financial interests or personal relationships that could have appeared to influence the work reported in this paper.